\documentclass[preprint,12pt]{elsarticle}

\usepackage{amssymb}
\usepackage{amsmath}
\usepackage{multirow}

\usepackage{hyperref}
\usepackage{algpseudocode}
\usepackage{algorithm}
\usepackage{mathtools}
\usepackage{subcaption}
\usepackage{rotating}
\usepackage[T1]{fontenc}
\usepackage{booktabs}
\usepackage{colortbl}
\usepackage{xcolor}
\usepackage{caption}

\newcommand{\subs}{S \in \mathcal{P} (M)}

\newcommand{\jointpost}{q_{\phi} (z|X)}
\newcommand{\qj}{q_{\phi_j} (z|x_j)}
\newcommand{\qgj}{q_{\phi_j} (z|g_j (x_j))}
\newcommand{\ljm}{\mathcal{L}_{uni}}
\newcommand{\numberthis}{\addtocounter{equation}{1}\tag{\theequation}}
\DeclareMathOperator{\argmax}{argmax}

\colorlet{tableheadcolor}{gray!25} 
\newcommand{\headcol}{\rowcolor{tableheadcolor}} %

\journal{Neural Networks}

\begin{document}

\begin{frontmatter}



\title{Bridging the inference gap in Mutimodal Variational Autoencoders}


\author[label1]{Agathe Senellart} 
\author[label1]{Stéphanie Allassonnière} 

\affiliation[label1]{organization={UMR1346, Universit\'e de Paris Cit\'e, Inria Paris, Inserm},
}

\begin{abstract}
    From medical diagnosis to autonomous vehicles, critical applications rely on the integration of multiple heterogeneous data modalities. Multimodal Variational Autoencoders offer versatile and scalable methods for generating unobserved modalities from observed ones. Recent models using mixtures-of-experts aggregation suffer from theoretically grounded limitations that restrict their generation quality on complex datasets. In this article, we propose a novel interpretable model able to learn both joint and conditional distributions without introducing mixture aggregation. Our model follows a multistage training process: first modeling the joint distribution with variational inference and then modeling the conditional distributions with Normalizing Flows to better approximate true posteriors. Importantly, we also propose to extract and leverage the information shared between modalities to improve the conditional coherence of generated samples. Our method achieves state-of-the-art results on several benchmark datasets. 
\end{abstract}


\begin{highlights}
\item Development of two novels methods for modeling and generating multimodal data, based on the Variational Autoencoder framework, Normalizing Flows and Self-Supervised Learning.
\item Comprehensive evaluation showing superior performance compared to state-of-the-art models on several benchmark datasets.
\end{highlights}


\end{frontmatter}



\section{Introduction}\label{sec1}

In many cases, information is conveyed through multiple heterogeneous modalities. In the medical field, a patient's status is comprehensively described through
various analyses: sonograms, MRI, blood analysis, textual reports,
 etc \ldots
Taking different views into account jointly leads to richer representations
and a better understanding of the modalities' interactions. 
Two important challenges in multimodal machine learning are the tasks of 
\textit{learning relevant joint representations} and
\textit{generating realistic data}, either \emph{from one modality to another or 
in all modalities jointly}. 

Multimodal Variational Autoencoders are latent generative models that can be used to tackle both challenges at the same time. 
In recent years, several approaches have been proposed to extend the Variational Autoencoder~\cite{kingma_auto-encoding_2014} (VAE) to efficiently model multimodal data. Some of them suggest training \emph{coordinated} VAEs where the latent spaces for all modalities are constrained to be similar~\cite{higgins_scan_2018,tian_contrastive_2020,yin_associate_2017}. In other works, a single latent space is used to jointly represent all modalities~\cite{suzuki_joint_2016,wu_multimodal_2018,shi_variational_2019}. Among these models, 
one popular approach is to aggregate modalities using a
simple function such as Product-of-Experts~\cite{wu_multimodal_2018}, or Mixture-of-Experts~\cite{shi_variational_2019},~\cite{sutter_generalized_2021}. Aggregation has the advantage of requiring fewer parameters and therefore being easily scalable. However, recent works show that it can limit the quality and diversity of generated samples~\cite{daunhawer_limitations_2022, sutter_generalized_2021}.

In this article, we propose a new flexible VAE-based framework that can model the
 \textbf{joint and conditional distributions across any number of modalities}.
    In particular, our main contributions are:
\begin{itemize}
    \item Development of two novel multimodal VAE-based methods for modeling and generating multimodal data. Unlike recent approaches, we do not use aggregation, which enables us to improve generation, particularly by integrating Normalizing Flows and leveraging shared information across modalities.
    \item Comprehensive evaluation on several benchmark datasets demonstrating that the proposed models outperform recent methods.
\end{itemize}

\section{Background}

Mathematically speaking, we assume that we observe multimodal samples $X = (x_1, x_2, \dots, x_M)$ with $M$ modalities from an 
unknown distribution $p(X)$. We aim to approximate this joint distribution as well as the conditional distributions with parametric ones $p_{\theta}(X)$,  $p_{\theta}(x_j|x_i)$ for any $1 \leq i \neq j \leq M$.
$p_{\theta}(x_j|x_i)$ is the distribution of one modality $x_j$ \emph{given} $x_i$.

In the VAE framework, one assumes that there exists a shared latent representation $z$, from which all modalities can be generated with parametric distributions ${( p_\theta(x_j|z) )}_{ 1 \leq j \leq M}$ called \emph{decoders}. For instance, for an image modality $x_1$, $p_{\theta}(x_1|z)$
can be a Gaussian distribution $\mathcal{N}(\mu_{\theta}(z), \Sigma_{\theta}(z))$ whose mean and variance are given by a neural network.  
 In most cases~\cite{wu_multimodal_2018,shi2021relating,suzuki_joint_2016}, each modality is supposed to be conditionally independent of the others given $z$, such that the joint model writes:
\begin{equation}\label{eq:independence}
    p_\theta(X,z) = p_{\theta}(X|z)p_{\theta}(z) =  p_{\theta}(z)\prod_{j=1}^{M} p_{\theta}(x_j|z)\,,
\end{equation}

where $p_{\theta}(z)$ is a \emph{prior} distribution over the latent variables and $\theta$ refers to all parameters used to model the prior and the decoders.
In that framework, the two goals mentioned above (model the joint and conditional distributions) translate as follows: first, we want to learn the best possible 
$\theta$  to model the observations. Secondly, we want to approximate the \emph{inference distributions} $p_{\theta}{(z| (x_j)_{j \in S})}$ 
to infer the latent variable from any given subset of modalities $\subs$ where $\mathcal{P}(M) = \{ S | S \subset [|1, M|] \text{ and } S \neq \emptyset \}$. If we can infer $z$ from observed modalities, we can then generate unobserved modalities with the decoders $(p_\theta(x_j|z))_{1 \leq j\leq M}$. In the rest of the article, we note $x_S \coloneqq (x_j)_{j \in S}$ to simplify notations. 

\subsection{Estimating the generative model's parameter $\theta$}
Given $N$ multimodal observations $(X^{(i)})_{1 \leq i \leq N}$, a natural objective to estimate $\theta$ is to optimize the log-likelihood of the data~\cite{kingma_auto-encoding_2014}:
\begin{equation*}
    \theta^* \in \underset{\theta}\argmax \sum_{i=1}^N \log p_\theta(X^{(i)})  =  \underset{\theta}\argmax \sum_{i=1}^N \left( \log \int_z p_\theta(X^{(i)},z)dz \right) \,.
\end{equation*}
Since this objective is intractable, one can resort to Variational Inference~\cite{jordan_introduction_1998,kingma_auto-encoding_2014} 
by introducing an auxiliary parametric distribution $q_{\phi}(z|X)$ allowing us to derive an unbiased 
estimate of the likelihood of the data:
\begin{equation}
    \widehat{p}_{\theta}(X,z) = \frac{p_{\theta}(X, z)}{q_{\phi}(z|X)} \hspace{5mm}\text{such that}\hspace{5mm} p_{\theta}(X) = \mathbb{E}_{q_{\phi}(z|X)}\left[\widehat{p}_{\theta}\right]\,.
\end{equation}
Then, using Jensen's inequality allows us to derive a lower bound on $\log p_{\theta}(X)$, referred to as the Evidence Lower Bound (ELBO). 
\begin{align*}
        \log p_{\theta}(X) &= \log \mathbb{E}_{q_{\phi}(z|X)} \left [ \widehat{p}_{\theta}\right ]\\
        &\geq \mathbb{E}_{q_{\phi}(z|X)}\left[\log  p_{\theta}(X|z) \right]  - KL(q_{\phi}(z|X) || p_{\theta}(z))
        = \mathcal{L}(X; \theta, \phi) \numberthis \label{elbo def}\,.
\end{align*}

This bound is tractable and can be optimized through Stochastic Gradient Descent \cite{kingma_auto-encoding_2014}. 
Noteworthy, the first term can be seen as a reconstruction error and the second as a regularization term 
encouraging latent embeddings to follow the prior distribution 
\cite{ghosh_variational_2020}. 
The distribution $q_{\phi}(z|X)$ is generally called the \textit{encoder} and one may prove that: 
\begin{equation}
    \mathcal{L}(X; \theta, \phi) = \log p_{\theta}(X) - KL(\jointpost || p_{\theta}(z|X))\,.
\end{equation}

 This implies that maximizing $\mathcal{L}(X; \theta, \phi)$ with respect to $\phi$ leads to minimizing the Kullback-Leibler (KL) divergence between the true posterior $p_{\theta}(z|X)$ and its variational approximation $\jointpost$ \cite{kingma_auto-encoding_2014}.
Some models also rely on variations of Eq.~\eqref{elbo def} to learn $\theta$: \cite{sutter_generalized_2021,palumbo_mmvae_2023} adds a $\beta$ factor to weigh the KL term in \eqref{elbo def}. That hyperparameter can be tuned to promote disentanglement in the latent space \cite{higgins_beta-vae_2017}: by increasing the KL term, it increases pressure on the latent variables to be independent, so that a single unit might encode a single generative factor. Other models \cite{shi_variational_2019, palumbo_mmvae_2023}
use a k-sampled importance weighted estimate of the log-likelihood (IWAE bound) \cite{burda_importance_2016} or replace the KL with a Jensen-Shannon divergence~\cite{sutter_multimodal_2020}. 

\subsection{Choice of the approximate inference distribution}

A simple choice is to model the approximate posterior $q_{\phi}(z|X)$ as a Gaussian distribution $\mathcal{N}(\mu_{\phi}(X), \Sigma_{\phi}(X))$ where a dedicated joint encoder network takes all modalities as input and outputs the parameters $\mu_{\phi}(X), \Sigma_{\phi}(X)$. By maximizing $\mathcal{L}$, we obtain an estimation of $\theta$ and an approximation of the joint posterior $p_{\theta}(z|X)$ with $q_{\phi}(z|X)$. 
However, we do not have access to the remaining subset posteriors $(p_{\theta}(z|x_S))_{\subs}$ which are \emph{intractable}. 
To estimate these posterior distributions, two approaches have been proposed, which we detail in the following paragraphs.

\subsection{Surrogate distributions and learning objectives}

First, a few models such as JMVAE \cite{suzuki_joint_2016}, or  TELBO \cite{vedantam_generative_2018} introduce surrogate parametric distributions $(q_{\phi_S}(z|x_S))_{\subs}$ and train them with an additional loss function to approximate the desired posterior distributions. However, those models use quite a large number of parameters since the joint posterior $q_{\phi}(z|X)$ and each approximate posterior $\left( q_{\phi_S}(z|x_S) \right)_{\subs}$ use a dedicated network encoder.
The number of parameters then scale with the number of subsets $|\mathcal{P}(M)| = 2^M$. 

\subsection{Aggregated models}

\emph{Aggregated models} compute the joint posterior $\jointpost$ as an aggregation of unimodal encoders
 $\qj$ for $ 1 \leq j \leq M$. 
MVAE \cite{wu_multimodal_2018} uses a Product-of-Experts (PoE) operation 
$\jointpost \propto p_{\theta}(z)\prod_j \qj$ while MMVAE \cite{shi_variational_2019} uses a Mixture-of-Experts (MoE). Many variants were then introduced such as Mixture-of-Product of Experts \cite{sutter_generalized_2021} or Generalized Product of Experts \cite{aguila_2023}. 
 Such a choice for $q_{\phi}(z|X)$ has several advantages. First it reduces the number of trainable parameters since $q_{\phi}(z|X)$ shares the same parameters as the unimodal encoders $(\qj)_{1 \leq j \leq M}$. To model a subset posterior $q_{\phi}(z|x_S)$ for $\subs$, no additional parameter is necessary; one can simply aggregate on the modalities in $S$. Therefore, these models are easily scalable to large number of modalities.
Furthermore, optimizing Eq.~\ref{elbo def} allows to optimize the generative parameter $\theta$ 
 and all inference parameters $\phi = (\phi_j)_{1\leq j \leq M}$ without introducing additional objectives to the loss function.
In particular, \cite{sutter_generalized_2021} rewrites the ELBO \eqref{elbo def} to explicitly highlight how 
 these aggregation methods encourage each estimated posteriors $q_{\phi_j}(z|x_j)$ to be close to the true joint posterior
  $p_{\theta}(z|X)$. 

However, it has been shown \cite{daunhawer_limitations_2022} that all \emph{mixture-based} models suffer from a fundamental limitation that caps their generative quality. 
More precisely, for these models, there is a generative discrepancy $\Delta(X)$ between the log-likelihood of the data and the ELBO:

\begin{equation}
    \label{discrepancy}
    \mathbb{E}_{p(X)}(\log(p_\theta(X))) \geq  \mathbb{E}_{p(X)}(\mathcal{L}(X; \theta, \phi)) + \Delta(X)\,,
\end{equation} 
where $p(X)$ is the observed empirical distribution.
$\Delta(X)$ is strictly positive and only depends on the law of $X$ and the mixture components~\cite{daunhawer_limitations_2022}.

Using $\mathbb{E}_{p(X)}(\log(p_\theta(X)) - \mathcal{L}(X; \theta, \phi)) =\mathbb{E}_{p(X)} \left( KL\left(q_{\phi}(z|X) || p_{\theta}(z|X)\right) \right)$, 
one can rewrite \eqref{discrepancy} as:

\begin{equation}
    \mathbb{E}_{p(X)} \left( KL(q_{\phi}(z|X) || p_{\theta}(z|X)) \right) \geq \Delta(X)\,.
    \label{mixture bound}
\end{equation}
This lower bound implies that the approximate joint posterior $\jointpost$ can only approach the true joint posterior $p_{\theta}(z|X)$ up to $\Delta(X) > 0$.

The authors in \cite{daunhawer_limitations_2022} detail in extensive experiments how these generative discrepancy results in a diminished quality of generated samples. 

For aggregated models that are only based on a Product-of-Experts such as the MVAE, this issue is avoided but a trade-off is 
observed between the generative coherence and the generative diversity \cite{sutter_generalized_2021}.

\subsection{Recent developments}

In order to compensate for this diversity/coherence trade-off, additional terms might be added to the ELBO to further ensure certain properties of the unimodal encoders. 
For instance, the MVTCAE model adds Conditional Variational Information Bottleneck (CVIB) terms to the ELBO
    \cite{hwang_multi-view_2021} while the CRMVAE model adds unimodal reconstruction terms \cite{suzuki2023mitigating}. Another approach is to modify the training paradigm with a contrastive learning objective \cite{shi2021relating}. Recently, methods have been proposed with more complex generative models including multiple, separated 
\cite{sutter_generalized_2021,Lee_2021_CVPR,daunhawer_self-supervised_2021} or hierarchical latent variables \cite{vasco_leveraging_2022, palumbo2024deep}.
 An additional goal of these models is to separate into different latent spaces the information
  shared across modalities from modality-specific factors. Models using multiple latent variables are sometimes sensitive to
   the \emph{shortcut} issue, referring to shared information leaking into the modality specific latent spaces. 
   Recently, MMVAE+ \cite{palumbo_mmvae_2023} was proposed with an amended ELBO loss and modalities' specific priors to limit that 
   phenomenon \cite{palumbo_mmvae_2023}. 
   However the MMVAE+ is still based on a mixture aggregation and therefore suffers from the intrinsic limitation mentioned above in Eq.~\eqref{mixture bound}, which we observe in our experiments.
   
Finally, recent work complement multimodal VAEs with diffusion models to improve generative quality \cite{MLD, palumbo2024deep}. 

\section{Proposed method}

To overcome the generative discrepancy gap observed in mixture-based models, we propose to disentangle the training of the joint generative model $p_{\theta}(X)$ and the approximation of the posteriors $p_{\theta}(z|x_j)$ for $1 \leq j \leq M$ in the same line of work as \cite{suzuki_joint_2016,vedantam_generative_2018}.Therefore our method consists of two separate steps:
\begin{itemize}
\item Train a Variational Autoencoder to learn the generative model $\theta$ as well as an approximation of the joint posterior $q_{\phi}(z|X)$.

\item For conditional generation, approximate the unimodal posteriors  with Normalizing Flows~\cite{pmlr-v37-rezende15} $q_{\phi_j}(z|x_j)$ for $1 \leq j \leq M$.
\end{itemize}

For the subset posteriors, we show that, for any $\subs$, we can approximate $p_{\theta}(z|x_S)$ with a Product-of-Experts $\prod_{j \in S} q_{\phi_j}{(z|x_j)}/ p_{\theta}{(z)}^{|S|-1}$. This way, no additional network needs to be trained and our framework scales for large numbers of modalities.
 Note that this Product-of-Experts is only used during inference \emph{after} the training and not in the optimization of the multimodal ELBO, which means that it doesn't suffer from the same limitations as PoE aggregated models. In the following subsections, we detail each step of our method, and then we introduce an improvement that leverages information shared across modalities.

\subsection{Step 1: Training the joint generative model}
\label{joint_training_step}

For learning the generative parameter $\theta$, we optimize the ELBO \eqref{elbo def} with a $\beta$ factor weighting the regularization term. We model the joint encoder $q_{\phi}(z | X)$ as a Gaussian distribution $\mathcal{N}(\mu_{\phi}(X), \Sigma_{\phi}(X))$, with $\mu_{\phi}(X)$ and $\Sigma_{\phi}(X)$ given by a neural network taking all modalities as inputs. This step is exactly similar to training a unimodal VAE, and every improvement that was proposed for the unimodal case could be easily adapted here.

\subsection{Step 2: Learning the posterior distributions}\label{step2}

Once the generative model is learned, we freeze the generative model $p_{\theta}(X|z)$ and the joint encoder $\jointpost$. For $ 1 \leq j \leq M$ we introduce a surrogate distribution $\qj$  to approximate the unimodal posterior $p_{\theta}(z|x_j)$ that is intractable. To fit these distributions, we minimize the following objective introduced in \cite{suzuki_joint_2016}.
\begin{equation}
    \label{ljm}
    \ljm(X; (\phi_j)_{1 \leq j \leq M}) = \sum_{j=0}^{M} KL(q_{\phi}(z|X) | q_{\phi_j}(z|x_j))\,.
\end{equation}
Intuitively, minimizing~\eqref{ljm}  encourages $\qj$ to cover all the relevant modes or support of the trained posterior $\jointpost$.
Since $\jointpost$ is frozen, minimizing Equation~\eqref{ljm} amounts to minimizing: 
 \begin{equation}\label{ljmtilde}
    \tilde{\ljm}(X; {(\phi_j)}_{1 \leq j \leq M})= - \sum_{j=0}^{M} \mathbb{E}_{\jointpost} \left( \log \qj  \right)\,.
 \end{equation}
For $ 1\leq j \leq M$, the expectation inside the sum can be estimated by sampling $z\sim \jointpost$. Equation~\eqref{ljmtilde} shows that during training, the unimodal encoders are \emph{informed} by the joint encoder: a latent variable $z$ is sampled from $\jointpost$ and then for each $1\leq j\leq M$,   $\log \qj$ is maximized.
\cite{vedantam_generative_2018, suzuki_joint_2016} and~\cite{hwang_multi-view_2021} provide interesting interpretations of this objective that we detail in \ref{interpretations}. In particular, one can prove that for any $1 \leq j \leq M$, optimizing~\eqref{ljm} brings $\qj$ close to an average distribution $q_{\phi}^{(avg)}(z|x_j) \coloneqq \mathbb{E}_{p({(x_i)}_{ i\neq j }|x_j)}(q_{\phi}(z|X))$.

This loss function is used in~\cite{suzuki_joint_2016}, but the JMVAE model suffers from poor coherence in certain use cases. One reason for this is the use of Gaussian distributions to model $\qj$ for $1 \leq j \leq M$, which lacks flexibility for approximating the true posteriors. 
 We transform these gaussian distributions using Normalizing Flows which allow us to better approximate complex distributions. Normalizing Flows are a powerful modeling tool that enables the modeling of complex, differentiable distributions \cite{pmlr-v37-rezende15}.
  A flow is an invertible smooth transformation $f$ that can be applied to an initial distribution to create a new one,
   such that if $Z$ is a random vector with density $q(z)$, then $Z' = f(Z)$ has a density given by:
\begin{equation}
    q'(z') = q(z) \left|\det \frac{\partial f^{-1}}{\partial z'}\right| = q(z) \left |\det \frac{\partial f}{\partial z}\right |^{-1}\,.
\end{equation}
Combining $K$ transformations $z_K = f_K \circ f_{K-1} \circ \dots \circ f_1(z_0)$ allows us to gain in complexity of the final distribution.

In our case, for each modality $1 \leq j \leq M$, we model the approximate posterior $\qj$ with the following log-density:

\begin{equation}
    \begin{split}
    \log \qj = \log q_{\phi_j}^{(0)}(z_0|x_j) - \sum_{k = 1}^{K} \log \left| \det \frac{\partial f_k^{(j)}}{\partial z_{k-1}}\right|\,,
    \end{split}\label{flots mvae}
\end{equation}
where $q_{\phi_j}^{(0)}(z_0|x_j)$ is a simple parametrized Gaussian distribution, the parameters of which are given by neural networks,
and $(f_k^{(j)})_{1 \leq k \leq K}$ are Masked Autoregressive Flows \cite{papamakarios_masked_2017}. In section \ref{toy dataset}, we illustrate that this expression allow us to approximate much more precisely the true unimodal posteriors. 
Because of the joint training of Normalizing Flows during this step, we refer to our model as JNF.

\subsection{Sampling from the subset posteriors}

Recall that one of our goals is to be able to infer the latent variable $z$ from \emph{any subset of modalities} $\subs$. Until now, we have estimated the joint posterior with $\jointpost$ and the unimodal posteriors with $\qj$ for any $j \in  [|1,M|]$. Using the same derivation as \cite{wu_multimodal_2018}, we prove that we can approximate any subset posterior using the trained unimodal encoders. 

Let $\subs$ and $x_S = (x_j)_{j \in S}$:
\begin{align}
    p_{\theta}(z|x_S) &= \frac{p_{\theta}(x_S|z)p_{\theta}(z)}{p_{\theta}(x_S)}
    = \frac{p_{\theta}(z) \prod_{j \in S}p_{\theta}(x_j | z)}{p_{\theta}(x_S)} 
    = \frac{p_{\theta}(z) \prod_{j \in S}\frac{p_{\theta}(x_j , z)}{p_{\theta}(z)}}{p_{\theta}(x_S)}\\ &= \frac{ \prod_{j \in S} p_{\theta}(z|x_j)p_{\theta}(x_j)}{p_{\theta}(z)^{|S|-1} p_{\theta}(x_S)}
    = \frac{1}{Z}\frac{\prod_{j \in S} p_{\theta}(z|x_j)}{p_{\theta}(z)^{|S|-1}}
    \approx \frac{1}{Z}\frac{\prod_{j \in S} \qj}{p_{\theta}(z)^{|S|-1}}\label{eq:poe}
\end{align}
where $\frac{1}{Z} = \frac{\prod_{j \in S}p_{\theta}(x_j)}{p_{\theta}(x_S)}$ is a normalizing constant. We use Equation~\eqref{eq:independence} in the second equality. To sample from this distribution at inference time, we use Hamiltonian Monte Carlo (HMC) sampling  \cite{noauthor_mcmc_2011,betancourt2018HMC}
that enables sampling from any distribution with a differentiable density function known up to a multiplicative constant. We recall the algorithm for HMC in \ref{app:hmc}.

\subsection{An improvement of our method leveraging shared information}\label{condition on shared}

Up until now, we have not made any assumption regarding the interactions between modalities $(x_1, x_2, \dots, x_M$). However, in many multimodal datasets, there is an amount of \emph{shared} semantic information between modalities. For instance, in the MNIST-SVHN dataset \cite{lecun_gradient-based_1998,netzer_reading_2011}, the shared semantic content is the digit present in both images. The background information is \emph{modality-specific} in the sense that it doesn't affect other modalities. To generate unobserved modalities, one would only need the shared semantic content and not the modality specific information.
Therefore, it seems relevant to try to extract and use this  \emph{shared} information.
Formally, let us assume that for any $1 \leq j \leq M$ we have a projector $g_j$ such that:
\begin{equation}\label{eq:summary statistic}
 \forall 1 \leq i \leq M, p_{\theta}(x_i|x_j)  = p_{\theta}(x_i|g_j(x_j))\,.
\end{equation}
Morally speaking, $g_j$ extracts the information shared across modalities while tuning out the modality specific information. Then we can write:
\begin{equation}
    p_{\theta}(x_i|g_j(x_j)) = \int_z p_{\theta}(x_i|z)p_{\theta}(z | g_j(x_j))dz
\end{equation}

That is, to generate modality $x_i$ from modality $x_j$, we can learn to approximate $p_{\theta}(z|g_j(x_j))$ which 
might be simpler than approximating $p_{\theta}(z|x_j)$ if we use relevant functions $(g_j)_{1 \leq j \leq M}$.

We propose an improvement of our method, in which we use \emph{pretrained} functions $(g_j)_{1 \leq j \leq M}$ to extract shared information across modalities and model the distributions $q_{\phi_j}(z|g_j(x_j))$ instead of $q_{\phi_j}(z|x_j)$. In that case, we model $q_{\phi_j}(z|g_j(x_j))$ with Normalizing Flows and use the adapted loss function for step 2 (Section~\ref{step2}):
\begin{equation}
    \label{ljm-bis}
    \ljm^{(shared)}(X; (\phi_j)_{1 \leq j \leq M}) = \sum_{j=0}^{M} KL(q_{\phi}(z|X) | q_{\phi_j}(z|g_j(x_j)))\,.
\end{equation}

\paragraph{Extracting information shared across modalities}

How can we learn relevant functions $(g_j)_{1 \leq j \leq M}$ that would verify Equation~\eqref{eq:summary statistic}?
Many methods have been proposed to extract information shared across modalities, and the best method might depend on the dataset, which is why this is a flexible component of our method. In our experiments, we tried two general methods: Deep Canonical Correlation Analysis (DCCA) \cite{andrew_deep_2013} and Contrastive Learning (CL) \cite{poklukar_geometric_2022,clip_2021,abid_contrastive_2019,tian_constrastive_multiview_coding}. In both cases, the projectors $(g_j)_{1 \leq j \leq M}$ are trained \emph{jointly} to learn \emph{similar} representations across modalities. For the projections $(g_j(x_j))_{1 \leq j \leq M}$ to be \emph{similar} across modalities, the projectors have to extract shared information while discarding unrelated information. The notion of similarity is defined differently in both methods: DCCA maximizes correlation between projections while CL optimizes cosine similarity. We detail each method in \ref{app:extract shared}.
 We \emph{conjecture} that using these methods, we can extract summary statistics $(g_j(x_j))_{1 \leq j \leq M}$ verifying Equation~\eqref{eq:summary statistic} and check this assumption in our experiments. Note that the projectors $(g_j)$ are trained \emph{before} training the VAE and that existing pretrained networks could also be used. 

\begin{figure}
    \centering
    \begin{subfigure}[b]{0.27\textwidth}
        \centering
        \includegraphics[width=\textwidth]{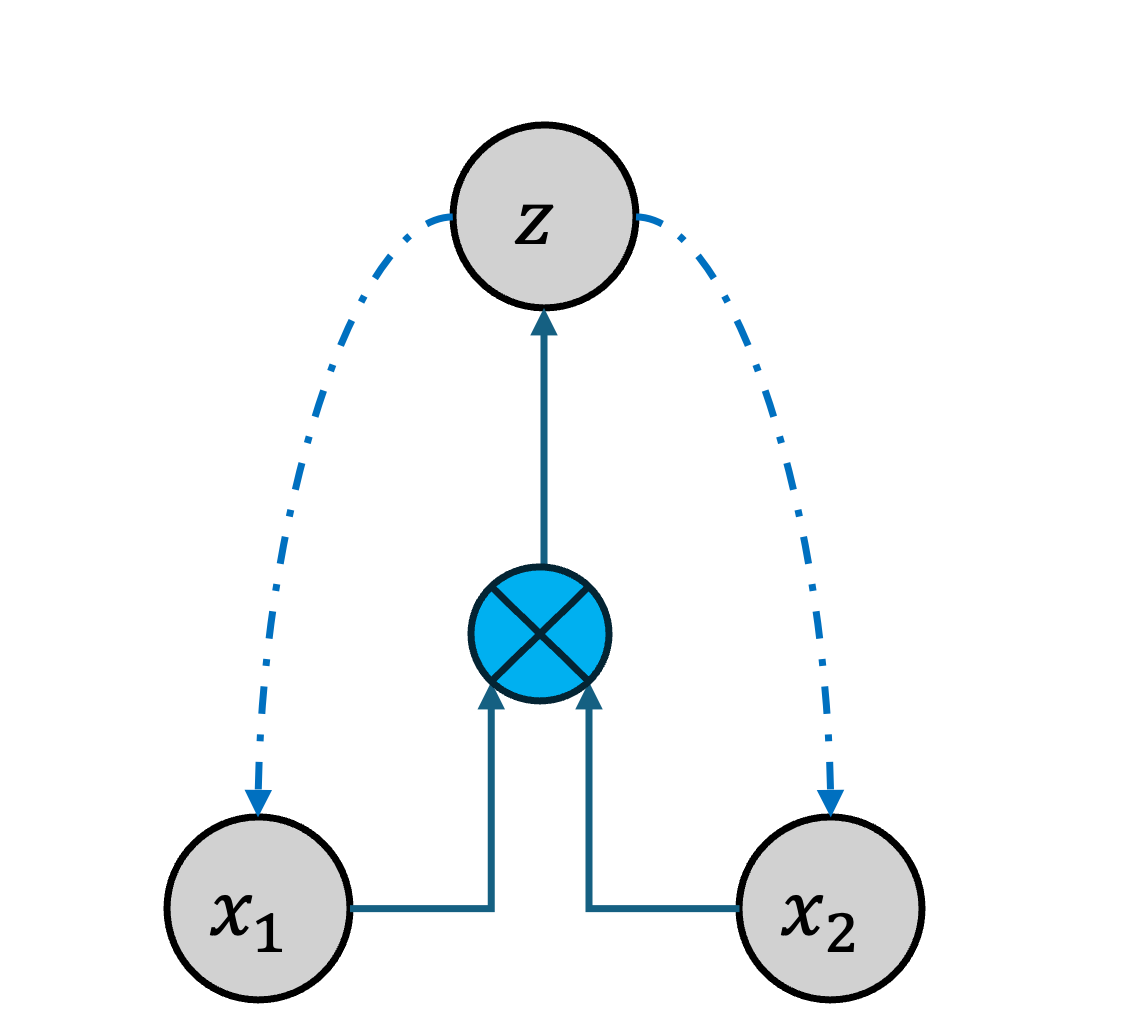}
        \caption{\tiny{Aggregated Models}}
    \end{subfigure}
    \hspace{0.5cm}
    \begin{subfigure}[b]{0.27\textwidth}
        \centering
        \includegraphics[width=\textwidth]{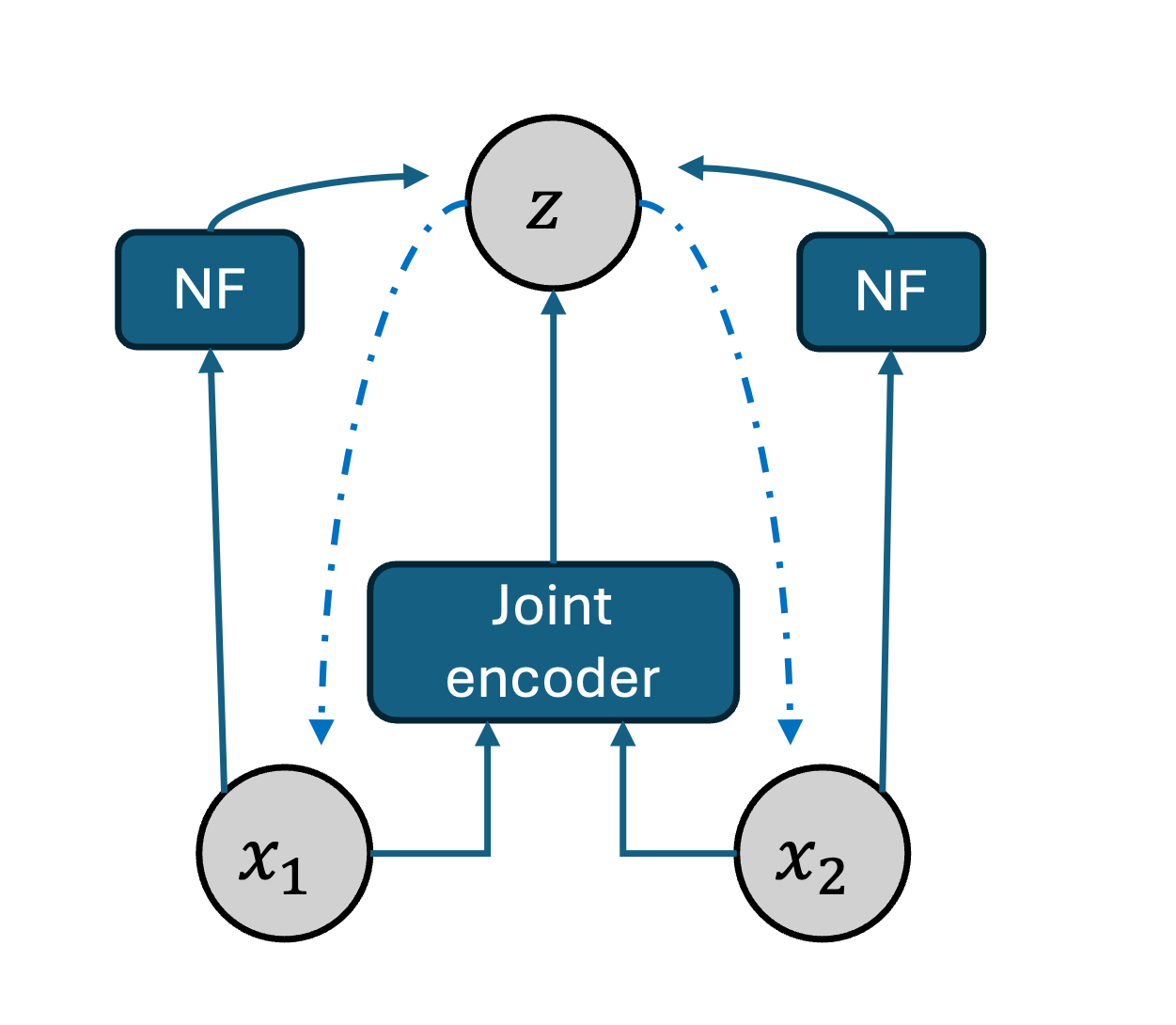}
        \caption{\tiny{JNF model}}
    \end{subfigure}
    \hspace{0.5cm}
    \begin{subfigure}[b]{0.27\textwidth}
        \centering
        \includegraphics[width=\textwidth]{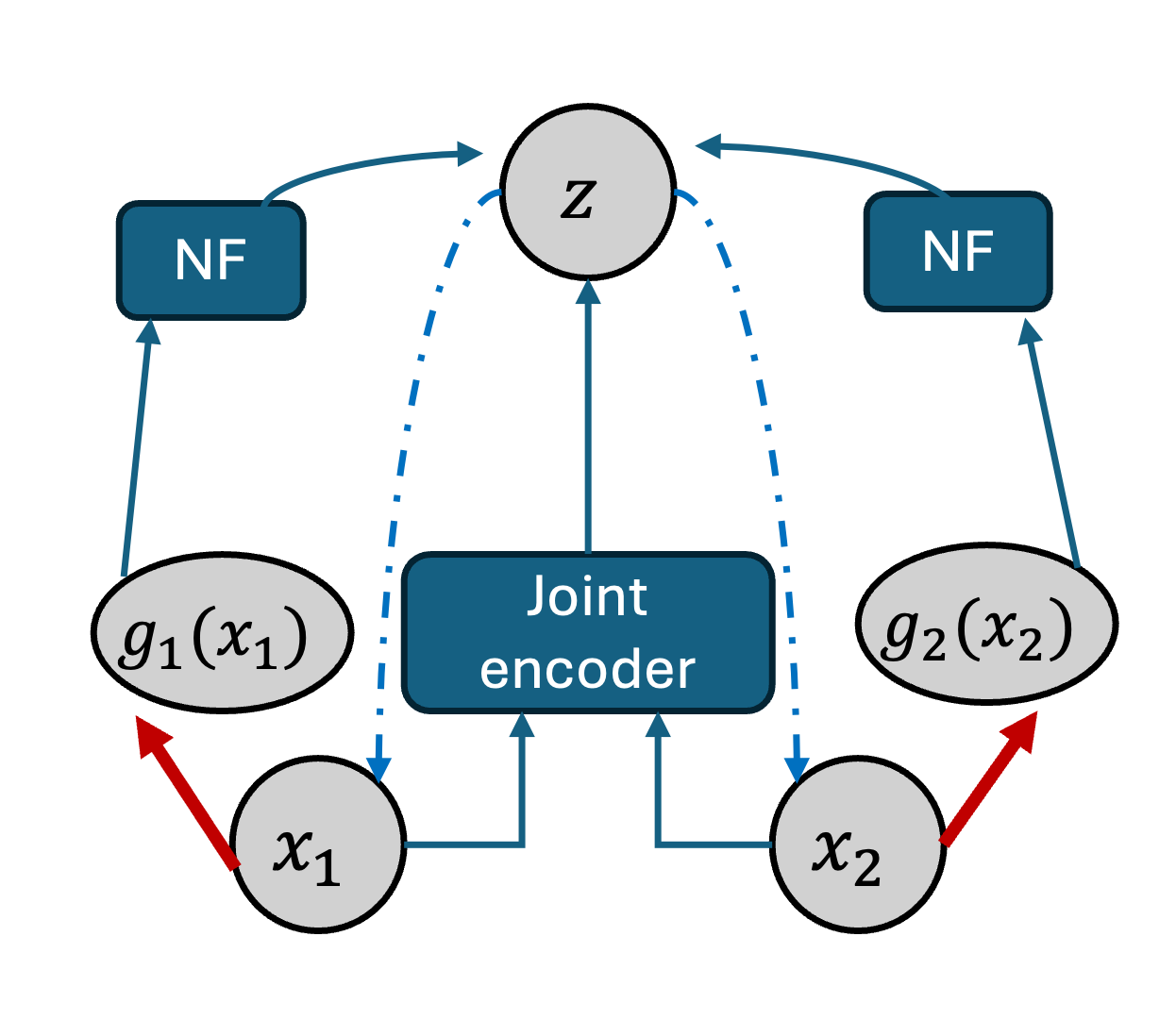}
        \caption{\tiny{JNF-Shared model}}
    \end{subfigure}
    \caption{Graphical models in the case $M=2$. Dashed lines represent decoders, solid lines represent encoders, and red arrows represent the projectors extracting shared information. "NF" refers to Normalizing Flows.}\label{fig:graphs}
\end{figure}

We refer to this improvement of our method as \emph{JNF-Shared}. In Figure~\ref{fig:graphs}, we summarize both models and aggregated models in the case $M=2$.

\section{Experiments}\label{sec : experiments}

In this section, we first illustrate our method on a toy dataset, and then compare results against state-of-the-art methods.

\subsection{Toy dataset}
\label{toy dataset}

\begin{figure}[htb]
    \includegraphics[width = \linewidth]{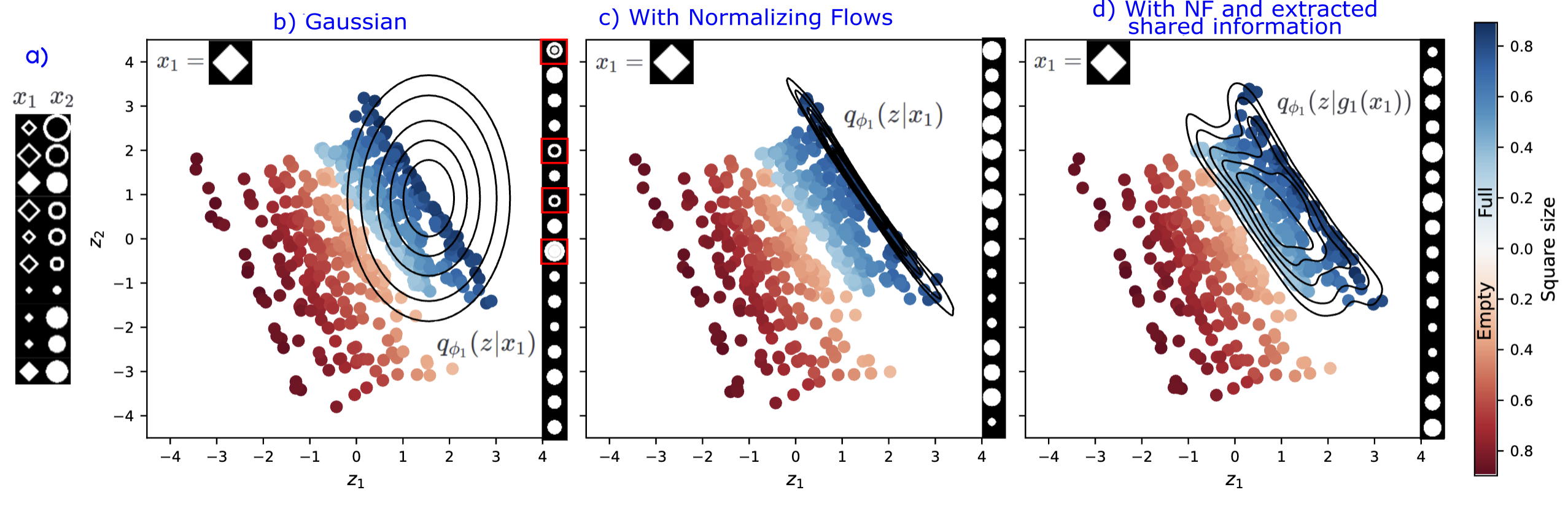}
    \caption{a) Samples from the toy dataset.
    b) The joint generative model $p_{\theta}(x_1,x_2)$ has been 
    learned and we visualize the 2-dimensional latent space. Each point encodes a 
    pair of images $(x_1,x_2)$. Here the color of each point, indicates the 
    size and class of the encoded \emph{square}. We try to approximate the posterior $p_{\theta}(z|x_1)$ of a large square image $x_1$ (shown in the top left), that
    corresponds to \emph{dark blue dots} in the latent space. In b), we use a diagonal Gaussian distribution
    and in c) we use Normalizing Flows. We see that Normalizing Flows capture
    a realistic posterior where the Gaussian distribution has a support that is too large, leading to unrealistic generation framed in red.
    d) Using DCCA, we extract the information shared across modalities, which is the shape class: full or empty. We learn $q_{\phi_1}(z|g_1(x_1))$ and see that it approximates well the part of the latent space which encodes full shapes. For b), c), and d)  
    we present samples generated in the circle modality using the learned posterior on the right side of each plot. Both c) and d) produce
    relevant and diverse samples.}
    \label{fig:toydata}
\end{figure}

We design a toy dataset with two black and white image modalities: 
$x_1$ is a square and $x_2$ is a circle. The sizes of each shape are independent. There are two classes of shapes: the \emph{full} shapes and the \emph{empty} ones.
This class is shared across modalities: if the circle is full, the square is full regardless of their size. Figure~\ref{fig:toydata} presents samples of this toy dataset. We perform the first step of our method on this dataset (see \ref{joint_training_step}), which is training a simple joint VAE with a two-dimensional latent space that we can visualize. In Figure~\ref{fig:toydata}, we can see how this joint latent space is structured,
with the full shapes on one side (blue dots) and the empty shapes on the other side (red dots). In Figure~\ref{fig:toydata} ,the intensities of the colors indicate the size distribution with larger squares encoded away from the center.

Using this well-structured latent space we then train $q_{\phi_1}(z|x_1)$ to approximate $p_{\theta}(z|x_1)$ using our objective $\ljm$~\eqref{ljm}. We display an example distribution $q_{\phi_1}(z|x_1)$ that we obtain for $x_1$ being a large full square. We compare results when modeling  $q_{\phi_1}(z|x_1)$ with either a Gaussian or Normalizing Flows (NF). The latter provides a more realistic approximation and generate coherent and diverse samples in the circles modality (shown on the right side of each plot).

In the last plot of Figure~\ref{fig:toydata}, we use the variant of our method where we first extract the information shared across modalities (here the emptiness or fullness of the shape) with a projector $g_1(x_1)$ and then approximate $q_{\phi_1}(z|g_1(x_1))$. Here, $g_1$ and $g_2$ are neural networks trained with the DCCA objective. We see in Figure~\ref{fig:toydata} right panel, that $q_{\phi_1}(z|g_1(x_1))$ covers well the part of the latent space corresponding to full samples and generates coherent and diverse samples. 
This shows that we have been able to capture both the conditional distribution and the shared information with $g_1(x_1)$ driving the conditional generation. On this toy dataset, both $p_{\theta}(z | x_1)$ and $p_{\theta}(z |g_1(x_1))$ are well approximated but on benchmark datasets, it appears that the latter is often easier to approximate than the former because it has a larger support. 

\subsection{Benchmark datasets and evaluation metrics}

We evaluate JNF and JNF-Shared on four benchmark datasets:

\begin{itemize}
    \item MNIST-SVHN introduced in~\cite{shi_variational_2019} that contains paired images from MNIST~\cite{lecun_gradient-based_1998} and the Street View House Numbers (SVHN) dataset~\cite{netzer_reading_2011}. The latter contains natural images of digits with diverse backgrounds and sometimes cropped distracting digits on the sides of the digit of interest.
    \item PolyMNIST introduced in \cite{sutter_generalized_2021} with five image modalities built from MNIST images with varied and complex backgrounds. This dataset allows to test the scalability of our method.
    \item Translated PolyMNIST introduced in \cite{daunhawer_limitations_2022} to demonstrate the limitations of mixture-based models. It is made of downscaled and translated digits with the same backgrounds as PolyMNIST. In \cite{daunhawer_limitations_2022} the authors point out that the generative performance is very degraded for mixture-based models on this dataset.
    \item Finally, we test our method on a dataset with heterogeneous modalities: the Multimodal Handwritten Digits dataset (MHD) \cite{vasco_leveraging_2022}
    which contains three modality types: image, sound and trajectory. 
\end{itemize}

We provide additional details and samples for each dataset in \ref{app:datasets}. We focus on conditional and unconditional generation and we evaluate:
\begin{itemize}
    \item the \emph{coherence} of multimodal samples. With pretrained classifiers, we assess whether the generated samples are consistent (i.e, share the same label) across modalities. 
    \item the \emph{diversity} of generated samples. To assess this diversity, we follow the procedures used in \cite{palumbo_mmvae_2023} and \cite{vasco_leveraging_2022}. For the 
    MNIST-SVHN and PolyMNIST datasets, we compute Fréchet Inception Distance \cite{heusel_gans_2017} (FID) between the distributions of true and generated samples. For the MHD dataset, the Inception network is not relevant to extract meaningful features since the modalities that we use are not natural images. 
    Therefore, we use pretrained, class-based and modality specific autoencoders to extract features for each sample and then compute the Mean Fréchet Distance (MFD)
    between true and generated samples. 
\end{itemize}

\subsection{Comparison to previous work}
We compare our method to several strong models: JMVAE \cite{suzuki_joint_2016}, MMVAE \cite{shi_variational_2019}, MoPoE \cite{sutter_generalized_2021}, MVTCAE model \cite{hwang_multi-view_2021}, MMVAE+ \cite{palumbo_mmvae_2023} and Nexus \cite{vasco_leveraging_2022}. We use implementations that were first validated by reproducing previous results. For a fair comparison, we use the same architectures and the same latent capacity across  models except for the MMVAE and MMVAE+ for which we use smaller latent spaces due to memory limitations from the K-sampled objective. We detail all hyperparameters in \ref{app:exp}. 
We train all models with a $\beta$-weighted ELBO and keep the $\beta \in \{ 0.5,1,2.5\}$ that maximizes average coherence for each model. Each experiment is repeated with four different seeds. We try training the projectors $(g_j)$ for our model JNF-Shared with Contrastive Learning (CL) or DCCA and report results for both.  

\subsection{Experimental results}
\begin{figure}[H]
    \centering
    \begin{subfigure}[b]{0.13\textwidth}
        \centering
        \includegraphics[width=\textwidth]{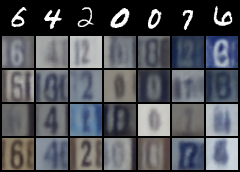}
    \end{subfigure}
    \hfill
    \begin{subfigure}[b]{0.13\textwidth}
        \centering
        \includegraphics[width=\textwidth]{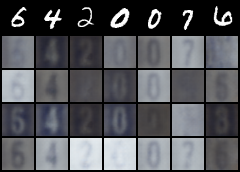}
    \end{subfigure}
    \hfill
    \begin{subfigure}[b]{0.13\textwidth}
        \centering
        \includegraphics[width=\textwidth]{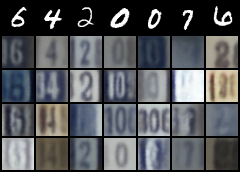}
    \end{subfigure}
    \hfill
    \begin{subfigure}[b]{0.13\textwidth}
        \centering
        \includegraphics[width=\textwidth]{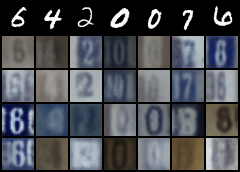}
    \end{subfigure}
    \hfill
    \begin{subfigure}[b]{0.13\textwidth}
        \centering
        \includegraphics[width=\textwidth]{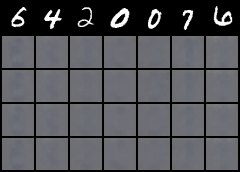}
    \end{subfigure}
    \hfill
    \begin{subfigure}[b]{0.13\textwidth}
        \centering
        \includegraphics[width=\textwidth]{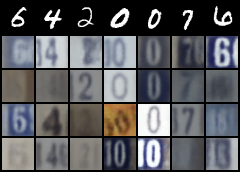}
    \end{subfigure}
    \hfill
    \begin{subfigure}[b]{0.13\textwidth}
        \centering
        \includegraphics[width=\textwidth]{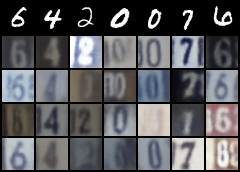}
    \end{subfigure}
    \vspace{2pt}
    \begin{subfigure}[b]{0.13\textwidth}
        \centering
        \includegraphics[width=\textwidth]{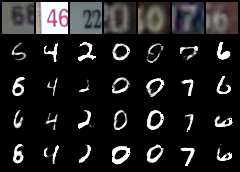}
    \end{subfigure}
    \hfill
    \begin{subfigure}[b]{0.13\textwidth}
        \centering
        \includegraphics[width=\textwidth]{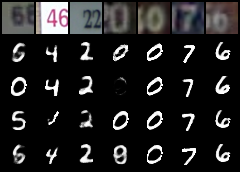}
    \end{subfigure}
    \hfill
    \begin{subfigure}[b]{0.13\textwidth}
        \centering
        \includegraphics[width=\textwidth]{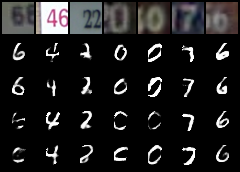}
    \end{subfigure}
    \hfill
    \begin{subfigure}[b]{0.13\textwidth}
        \centering
        \includegraphics[width=\textwidth]{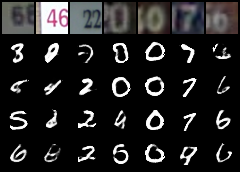}
    \end{subfigure}
    \hfill
    \begin{subfigure}[b]{0.13\textwidth}
        \centering
        \includegraphics[width=\textwidth]{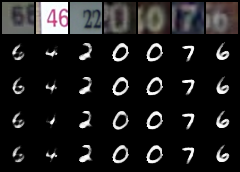}
    \end{subfigure}
    \hfill
    \begin{subfigure}[b]{0.13\textwidth}
        \centering
        \includegraphics[width=\textwidth]{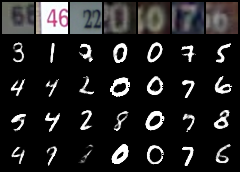}
    \end{subfigure}
    \hfill
    \begin{subfigure}[b]{0.13\textwidth}
        \centering
        \includegraphics[width=\textwidth]{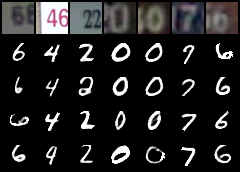}
    \end{subfigure}
    \begin{subfigure}[b]{0.13\textwidth}
        \centering
        \includegraphics[width=\textwidth]{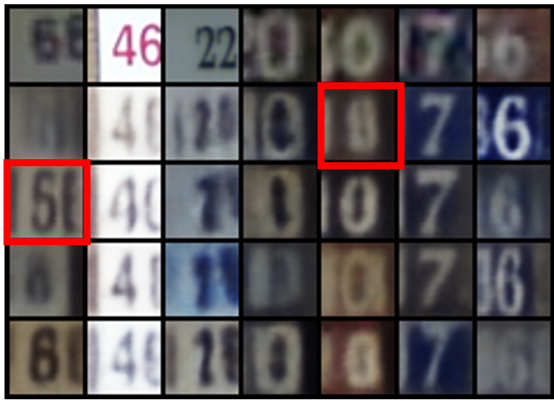}
        \caption{\tiny{JMVAE}}
    \end{subfigure}
    \hfill
    \begin{subfigure}[b]{0.13\textwidth}
        \centering
        \includegraphics[width=\textwidth]{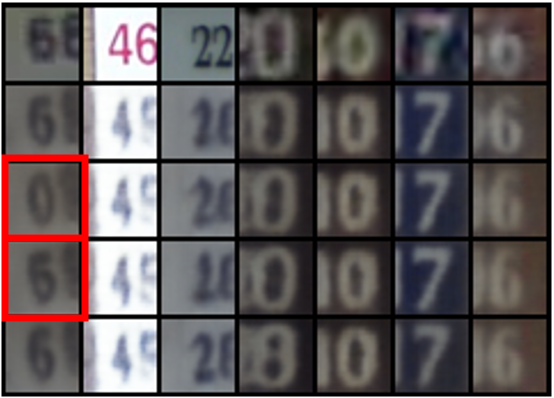}
        \caption{\tiny{MMVAE}}
    \end{subfigure}
    \hfill
    \begin{subfigure}[b]{0.13\textwidth}
        \centering
        \includegraphics[width=\textwidth]{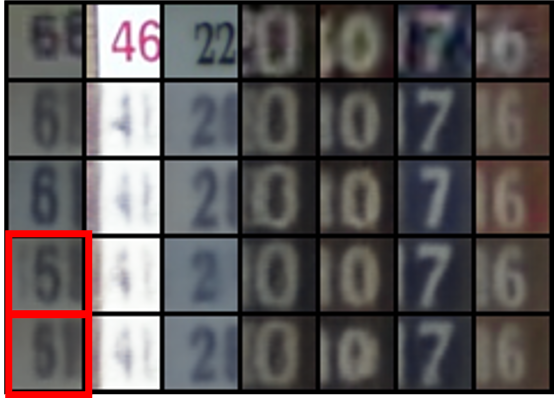}
        \caption{\tiny{MMVAE+}}
    \end{subfigure}
    \hfill
    \begin{subfigure}[b]{0.13\textwidth}
        \centering
        \includegraphics[width=\textwidth]{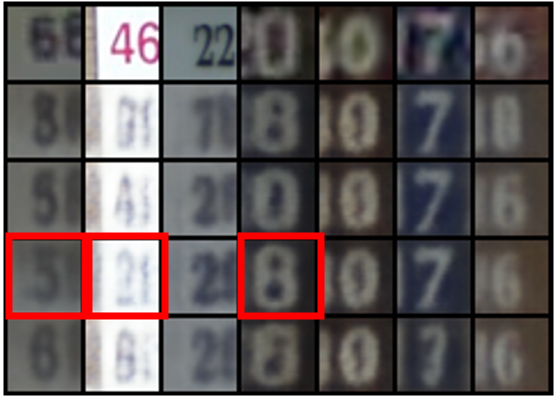}
        \caption{\tiny{MVTCAE}}
    \end{subfigure}
    \hfill
    \begin{subfigure}[b]{0.13\textwidth}
        \centering
        \includegraphics[width=\textwidth]{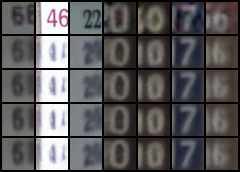}
        \caption{\tiny{MoPoE}}
    \end{subfigure}
    \hfill
    \begin{subfigure}[b]{0.13\textwidth}
        \centering
        \includegraphics[width=\textwidth]{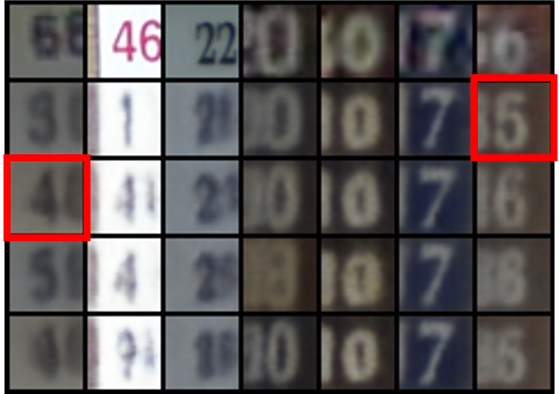}
        \caption{\tiny{JNF}}
    \end{subfigure}
    \hfill
    \begin{subfigure}[b]{0.13\textwidth}
        \centering
        \includegraphics[width=\textwidth]{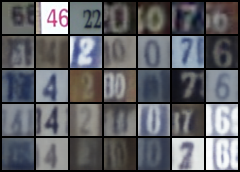}
        \caption{\tiny{JNF-CL}}
    \end{subfigure}
       \caption{On the first row: generation from MNIST to SVHN. On the second row: generation from SVHN to MNIST. On the third row: generation from SVHN to SVHN (unimodal reconstruction). In red, we frame samples where the background is well reconstructed but not the digit. JNF-CL refers to our model JNF-Shared with CL. Note that for this model, when reconstructing SVHN, we sample $z \sim q_{\phi_{2}}(z|g_{2}(x_{2}))$ and therefore the background information is filtered by the projector $g_{2}(x_{2})$ and cannot be reconstructed. However, the digit is well preserved which is what is required for cross-modal generation. }
       \label{fig:svhn to mnist}
\end{figure}

In Figure~\ref{fig:svhn to mnist},  we present generated samples and in Table~\ref{tab:ms_results}, we present quantitative results for the MNIST-SVHN dataset.

\begin{table}[htb]
    \footnotesize
    \vskip 0.15in
    \begin{center}
    \begin{tabular}{@{}lllll@{}}
    \toprule
    \bf{Model} & \bf{Joint} & \bf{M $\longrightarrow$ S} & \bf{S $\longrightarrow$ M}  & \bf{FID ($\downarrow$)} \\
    \midrule
    JMVAE &$\underline{0.43}\pm 0.10$ & $0.73\pm0.07$& $0.53\pm0.05$&  $57 \pm 3$ \\
    MMVAE $(k=10,\beta = 0.5)$&$0.35 \pm 0.02$  & $\underline{0.80}\pm0.01$ &$0.70\pm0.01$ &  $130 \pm 5$ \\
    MVTCAE &$\underline{0.44} \pm 0.02$& $\mathbf{0.81}\pm0.01$ & $0.52\pm0.02$ & $\mathbf{48}\pm 2$ \\
    MoPoE & $0.36\pm 0.01$& $0.12\pm 0.01$ & $\underline{0.72}\pm0.01$ & $359 \pm 12 $ \\
    MMVAE+ $(k=10)$& $\underline{0.43} \pm 0.05$ & $0.60 \pm 0.09$& $ 0.58\pm0.04$ &  $63\pm 5$\\
    \midrule
    JNF (Ours) & $ \mathbf{0.51} \pm 0.01$ & $\mathbf{0.82} \pm 0.01$ & $0.52 \pm 0.01$ &  $\underline{54}\pm 2$    \\
    JNF-Shared (DCCA) (Ours)& $\mathbf{0.51} \pm 0.01$ & $0.75 \pm 0.03$ & $0.69 \pm 0.05$ & $\underline{53} \pm 2$\\
    JNF-Shared (CL) (Ours)& $\mathbf{0.51} \pm 0.02$&$\mathbf{0.81} \pm 0.01$& $\mathbf{0.75}\pm0.02$ & $\mathbf{49}\pm 1$\\
    \bottomrule
    \end{tabular}
    \caption{Results on MNIST-SVHN. We present coherence for joint generation, conditional generation from MNIST (noted as $M$) to SVHN (noted as $S$) and vice-versa. FID values are computed on 50,000 SVHN images generated from MNIST. Best values are in bold and second-best are underlined.}\label{tab:ms_results}
    \end{center}
    \vskip -0.2in
\end{table}

Most models (except MoPoE and JNF-Shared) struggle to generate coherent MNIST images from SVHN images. We interpret this phenomenon by looking at reconstructed SVHN images in Figure~\ref{fig:svhn to mnist}. For many models, the background is well reconstructed but not the digit which is not well inferred using $q_{\phi_2}(z|x_2)$ (where $x_2$ is the SVHN modality). With JNF-Shared, the background is tuned out by the projector $g_2$ and the digit information is therefore better preserved when sampling $z \sim q_{\phi_2}(z|g_{2}(x_2))$. Our model JNF-Shared (CL) is the only one to reach competitive values \emph{for all metrics} on this dataset with coherent and diverse generations.

\begin{figure}[ht]
    \centering
    \begin{subfigure}[b]{0.19\textwidth}
        \centering
        \includegraphics[width=\textwidth]{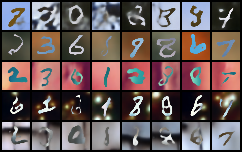}
        \caption{\scriptsize{JMVAE}}
    \end{subfigure}
    \begin{subfigure}[b]{0.19\textwidth}
        \centering
        \includegraphics[width=\textwidth]{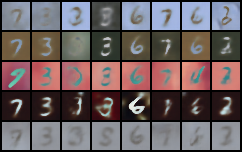}
        \caption{\scriptsize{MMVAE}}
    \end{subfigure}
    \begin{subfigure}[b]{0.19\textwidth}
        \centering
        \includegraphics[width=\textwidth]{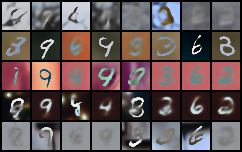}
        \caption{\scriptsize{MoPoE}}
    \end{subfigure}
    \begin{subfigure}[b]{0.19\textwidth}
        \centering
        \includegraphics[width=\textwidth]{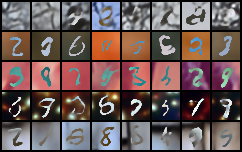}
        \caption{\scriptsize{MVTCAE}}
    \end{subfigure}
    \begin{subfigure}[b]{0.19\textwidth}
        \centering
        \includegraphics[width=\textwidth]{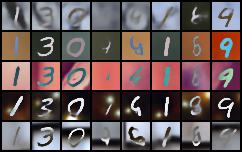}
        \caption{\scriptsize{MMVAE+}}
    \end{subfigure}
    \begin{subfigure}[b]{0.19\textwidth}
        \centering
        \includegraphics[width=\textwidth]{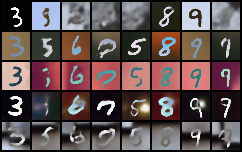}
        \caption{\scriptsize{JNF}}
    \end{subfigure}
    \begin{subfigure}[b]{0.19\textwidth}
        \centering
        \includegraphics[width=\textwidth]{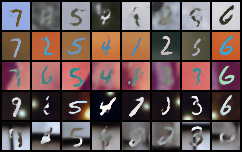}
        \caption{\scriptsize{JNF-CL}}
    \end{subfigure}
    \begin{subfigure}[b]{0.19\textwidth}
        \centering
        \includegraphics[width=\textwidth]{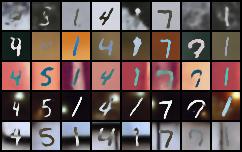}
        \caption{\scriptsize{JNF-DCCA}}
    \end{subfigure}
    \caption{Joint generation in all five modalities when sampling a latent code from the prior. In each image, each row corresponds to a modality. JNF-CL (resp. DCCA) correspond to our method JNF-Shared with CL (resp. DCCA).}\label{fig:mmnist unconditional}
\end{figure}

  JNF-Shared with CL projectors achieve higher coherence than DCCA projectors, which means that CL better extracts the shared information on this dataset. The MMVAE and MoPoE both produce SVHN samples that look 'averaged' resulting from the quality gap analyzed in \cite{daunhawer_limitations_2022}.
\begin{figure}[h!]
    \centering
    \begin{subfigure}[b]{\linewidth}
    \includegraphics[width=\textwidth]{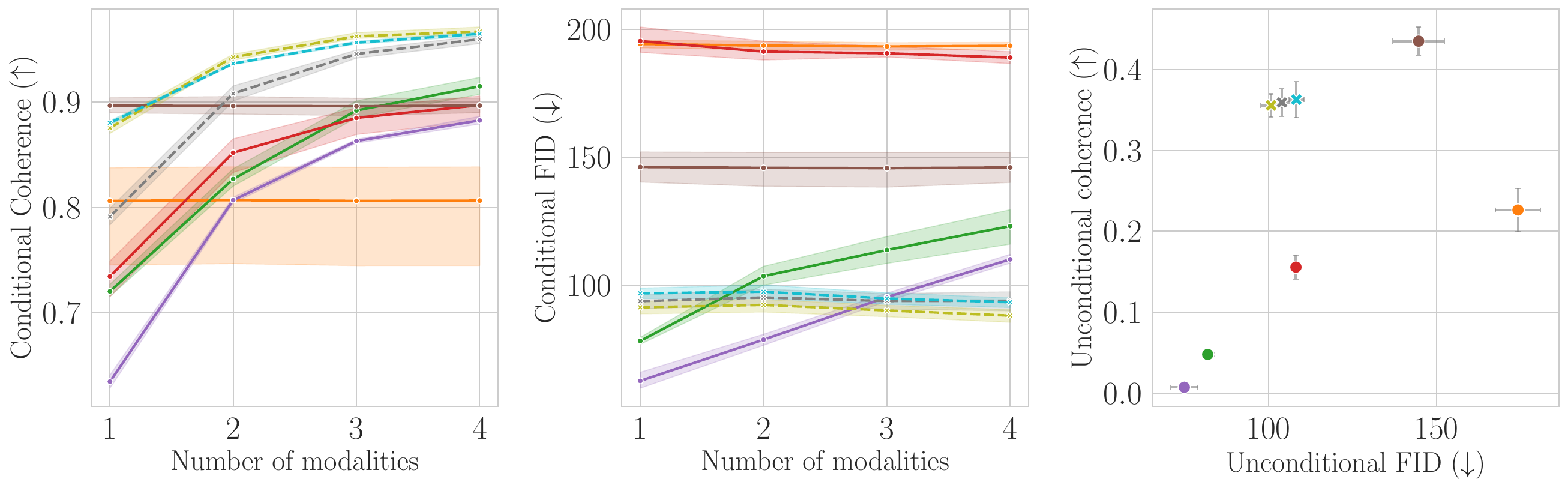}
    \caption{PolyMNIST}
    \end{subfigure}
    \begin{subfigure}[b]{\linewidth}
        \includegraphics[width=\textwidth]{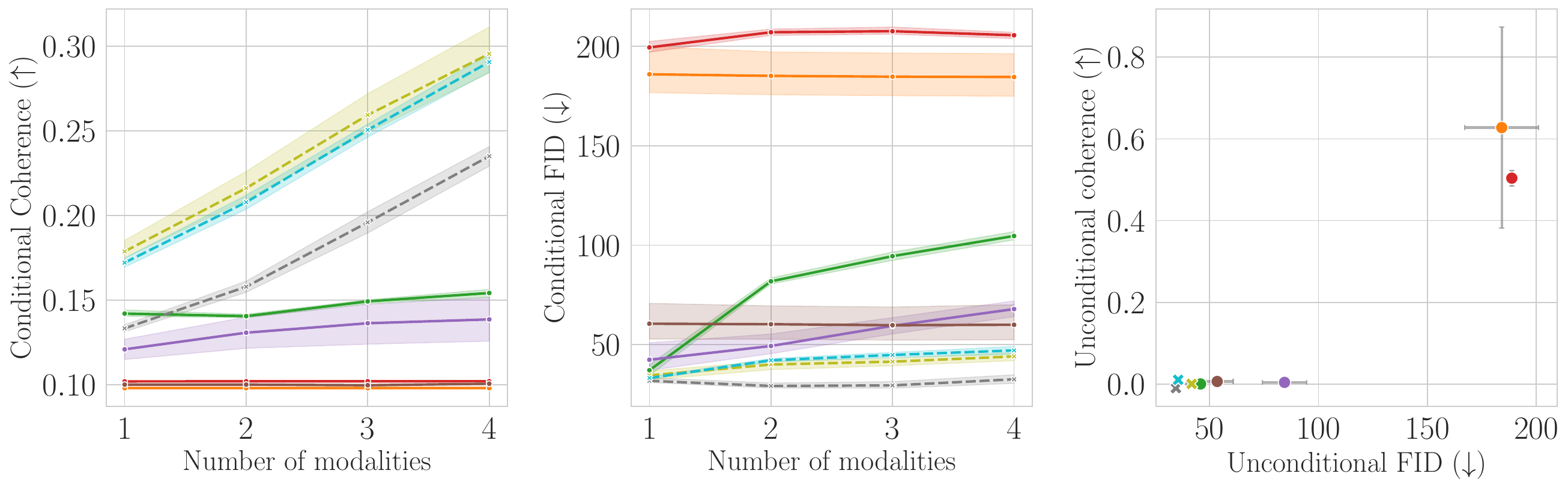}
        \caption{Translated PolyMNIST}
    \end{subfigure}
    \begin{subfigure}[b]{\linewidth}
        \centering
        \includegraphics[width=0.9\textwidth]{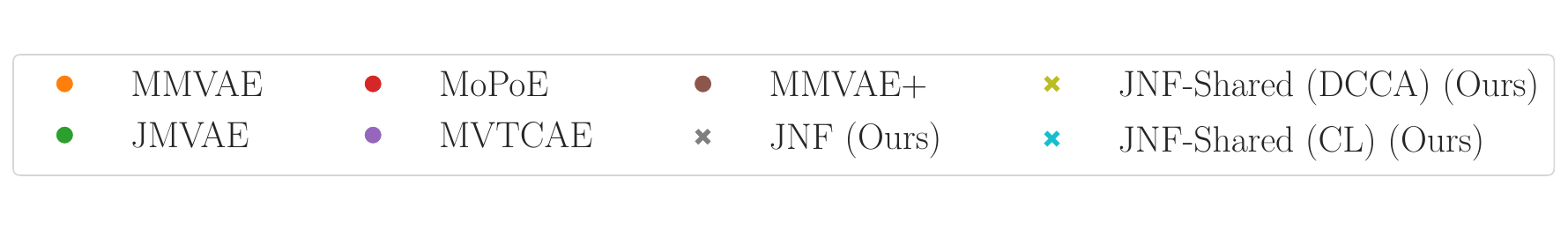}
    \end{subfigure}
    \caption{In the two left columns, we present results for conditional generation when varying the number of conditioning modalities. In the right column, we display coherence and FID for unconditional generation. Each point correspond to a different training seed. For these plots, best models having high coherence and low FID are in the top left corner. The FID is computed on 10,000 samples of the first modality.}\label{fig:mmnist results}
\end{figure}
In Figure~\ref{fig:mmnist results}, we present coherence and diversity results for all models on PolyMNIST and Translated PolyMNIST. We observe that our models reach the best coherence while maintaining low FID values. We present samples of unconditional generation in Figure~\ref{fig:mmnist unconditional}: our models produce coherent and diverse samples.
Our method JNF-Shared works well with both CL and DCCA projectors on this dataset.
In \cite{daunhawer_limitations_2022}, the authors observed that MMVAE and MoPoE show very degraded coherence on TranslatedPolyMNIST. We extend their observations to the MMVAE+ model that also has a conditional coherence close to 0.10, corresponding to random digit association. This is a direct consequence of the mixture aggregation, which limits generative quality on complex datasets \cite{daunhawer_limitations_2022}. In that setting, these models fail to extract any shared information across modalities. All models fail on the unconditional generation task from the prior: our models have good FID values but very low coherence. To improve this, a possible direction would be to fit a distribution on the latent embeddings \emph{after} training rather than sampling from the prior~\cite{ghosh2020}. On the contrary, MMVAE and MoPoE have high joint coherence but looking at the generated samples in~\ref{app:additional results}, we see that they only produce averaged images of the first digit.

In Table~\ref{fig:mhd_results}, we present results on the MHD dataset, where our models reach the best results for coherence and second best for diversity. For all datasets, additional results and generated samples can be found in \ref{app:additional results}.

\begin{table}[h]
    \footnotesize
    \centering
\begin{tabular}{@{}lllll@{}}
    \toprule
    & \multicolumn{2}{c}{\bf{Coherence $(\uparrow)$}} & \multicolumn{2}{c}{\bf{MFD $(\downarrow)$}}\\
    \midrule
     &Joint & Conditional & Joint & Conditional \\
    \midrule
    JMVAE  & $0.57 \pm 0.02$ & $0.86 \pm 0.01$ & $\mathbf{1.32} \pm 0.01$ & $0.29 \pm 0.02$ \\
    MMVAE & $\underline{0.63} \pm 0.01$ & $0.86 \pm 0.01$ & $1.63 \pm 0.05$ & $0.76 \pm 0.01$ \\
    MMVAE+ & $0.57 \pm 0.01$ & $\underline{0.89} \pm 0.01$ & $1.58 \pm 0.07$ & $0.55 \pm 0.08$ \\
    MVTCAE & $0.38 \pm 0.01$ & $0.87 \pm 0.01$ & $\bf{1.31} \pm 0.02$ & $\bf{0.13} \pm 0.01$ \\
    MoPoE  & $0.44 \pm 0.02$ & $0.74 \pm 0.01$ & $1.56 \pm 0.03$ & $2.17 \pm 0.03$ \\
    Nexus & $0.13 \pm 0.01$ & $0.34 \pm 0.01$ & $2.98 \pm 0.04$ & $3.36 \pm 0.03$ \\
    \midrule
    JNF(Ours)  & $\mathbf{0.67} \pm 0.01$ & $\underline{0.89} \pm 0.01$ & $\mathbf{1.32} \pm 0.02$ & $\underline{0.23} \pm 0.02$ \\
    JNF-Shared(CL)(Ours) & $\mathbf{0.65} \pm 0.02$ & $\mathbf{0.93} \pm 0.01$ & $\underline{1.35} \pm 0.04$ & $\underline{0.21} \pm 0.03$ \\
    JNF-Shared(DCCA)(Ours)& $\mathbf{0.66} \pm 0.01$ & $\mathbf{0.92} \pm 0.01$ & $\underline{1.37} \pm 0.04 $& $\underline{0.23} \pm 0.03$ \\
    \bottomrule
\end{tabular}
\caption{Experimental results on the MHD dataset. We present average coherence and MFD results for each model, for conditional and unconditional generation. Best values are in bold and second-best values are underlined.}\label{fig:mhd_results}
\end{table}

\section{Discussion and Perspectives}

In this article, we presented two novel VAE-based multimodal approaches for modeling and generating multimodal data. Several components of our methods are flexible and can be adapted to the use-case. For instance, the first step of our method consists of training a basic Joint Variational Autoencoder. However, many enhancements of the VAE have been  proposed to better learn the generative parameter $\theta$ with more expressive modeling of the posterior or prior distributions(\cite{rezende2016nf,pmlr-v84-tomczak18a,kingma_improving_2017}) or increased tightness of the objective bound function \cite{burda_importance_2016,tucker_dreg}. These improvements can be used in our framework to enhance the estimation of the joint generative model on which the rest of the model depends. We also introduced the idea of learning unimodal posteriors conditioned on \emph{a summary statistic} containing the information shared across modalities. For extracting the shared information, one can rely on Contrastive Learning or DCCA but also on other methods suited to the dataset. For instance Kernel Canonical Correlation Analysis~\cite{hardoon_canonical_2004} was used on functional imaging datasets \cite{bilenko_pyrcca_2016} or genetics \cite{ALAM201870}.
Finally, diffusion decoders \cite{Preechakul_2022_CVPR} could also be used to improve the quality of generated samples as was done in \cite{palumbo2024deep}.

\newpage

\appendix
\section{Details on the datasets used in the experiments}
\label{app:datasets}

\subsection{The MNIST-SVHN dataset}

To create this dataset, we paired images from the MNIST dataset \cite{lecun_gradient-based_1998} and the SVHN dataset \cite{netzer_reading_2011}. Previous work \cite{shi_variational_2019} paired each image in MNIST with 30 different images in SVHN to create a train set of 1 682 040 samples. To create a more challenging and realistic dataset, we only paired each image 5 times to have a smaller (yet still large) training dataset of 280 340 samples.

\subsection{PolyMNIST and Translated PolyMNIST Dataset}

\begin{figure}[H]
    \centering
    \begin{subfigure}[b]{0.4\textwidth}
        \centering
        \includegraphics[width = 0.7\textwidth]{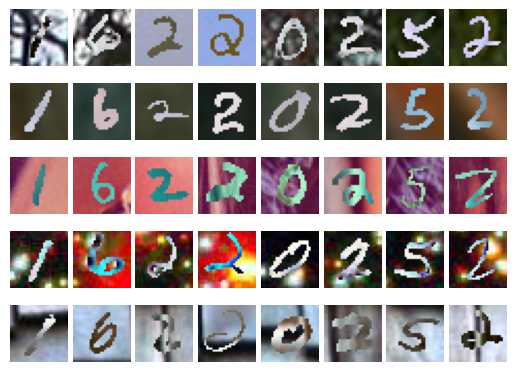}
        \caption{PolyMNIST.}
    \end{subfigure}
    \begin{subfigure}[b]{0.4\textwidth}
        \centering
        \includegraphics[width = 0.7\textwidth]{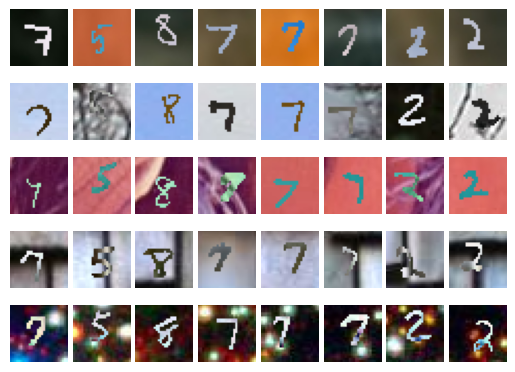}
        \caption{Translated PolyMNIST.}
    \end{subfigure}
    \caption{Eight multimodal samples for the PolyMNIST and TranslatedPolyMNIST dataset: each row correspond to a modality.}\label{datasets examples}
\end{figure}

In Figure~\ref{datasets examples}, we plot example images of the PolyMNIST and Translated PolyMNIST dataset used in the experiments
in section \ref{sec : experiments}. For the Translated PolyMNIST dataset, we downscale the digit by a factor 0.75 and add a random translation.
Each dataset contains 60 000 training samples and 10 000 test samples.

\subsection{Multimodal Handwritten Dataset}

\begin{figure}[H]
    \centering
    \includegraphics[width=0.6\linewidth]{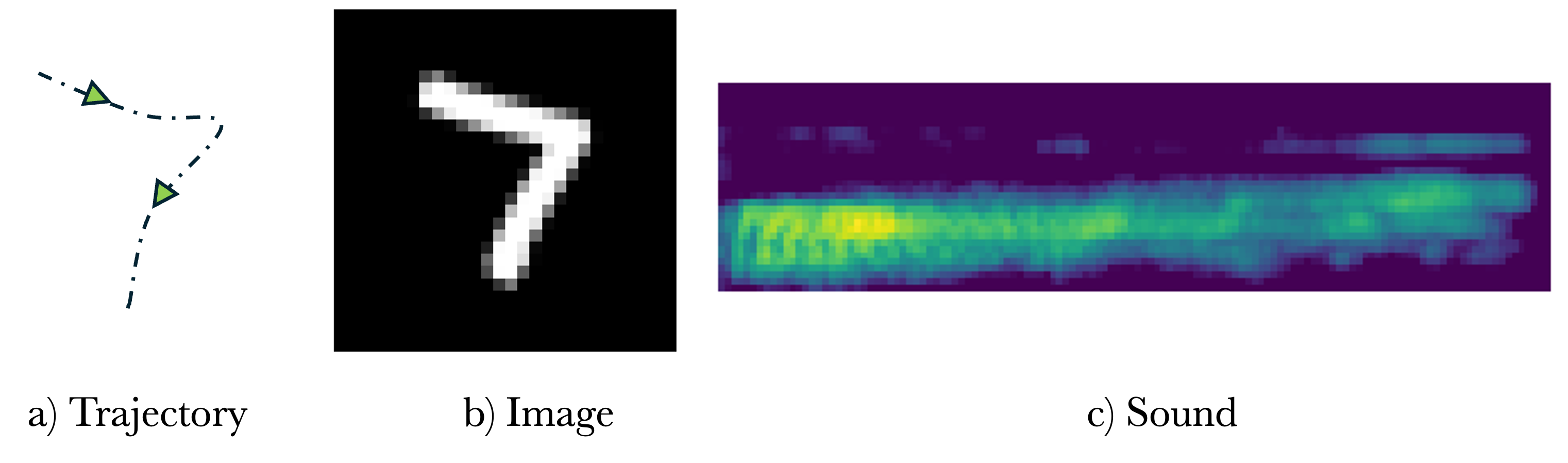}
    \caption{The MHD dataset that we use contains three modalities.}
\end{figure}

The 'Multimodal Handwritten Digits' (MHD) introduced in \cite{vasco_leveraging_2022} contains 4 modalities (including label): 
\begin{itemize}
    \item Image: gray digit images of size (1,28,28)
    \item Audio: spectrograms images with shape (1,32,128)
    \item Trajectory: flat arrays with 200 values
    \item Label : 10 categorical values
\end{itemize}

In our experiments, we don't use the label as a modality to make the task more challenging. 
This dataset contains 50 000 samples for training and 10 000 for testing.

\subsection{Toy dataset with circles and squares}

The images of circles and squares used in the toy experiment are of size (32,32) with black and white pixels.
All circles and squares are centered in the middle of the image with a minimum width of 10 pixels and a maximum width of 28 pixels.
This dataset contains 200,000 pairs of circles and squares. Half are empty and half are full.

\section{Methods to learn shared information across multiple modalities}\label{app:extract shared}

Here we detail two methods we have used to train the projectors $(g_j)_{1\leq j\leq M}$ to extract information \emph{shared} across modalities.
The projectors $(g_j)_{1\leq j\leq M}$ are trained \emph{before} training our multimodal VAE JNF-Shared that uses them.

\subsection{Deep Canonical Correlation Analysis}\label{dcca}

Deep Canonical Correlation Analysis \cite{andrew_deep_2013} (DCCA) aims at finding correlated neural representations for two complex modalities such as images. 
It is based upon the classical Canonical Correlation Analysis (CCA)~\cite{hardoon_canonical_2004} which we briefly recall here. Let $(X_1, X_2) \in \mathbb{R}^{n_1} \times \mathbb{R}^{n_2}$ two random vectors, $\Sigma_{1}, \Sigma_{2}$ their covariances matrices and $\Sigma_{1,2} = \mathrm{Cov}(X_1,X_2)$. CCA's objective is to find linear projections  $a^TX_1$, $b^TX_2$ that are maximally correlated :
\begin{equation*}
(a^*, b^*)  = \underset{a^T\Sigma_{1}a = b^T\Sigma_{2}b= 1}{\arg\max} a^T \Sigma_{1,2}b \,.
\end{equation*}
Once these optimal projections are found, we can set $(a_1,b_1) = (a^*, b^*)$ and search for subsequent projections $(a_i, b_i)_{2 \leq i \leq k}$ with the additional constraint that they must be uncorrelated with the previous ones. 
We can rewrite the problem of finding the first $k$ optimal pairs of projection as finding matrices $A \in \mathbb{R}^{(n_1,k)}$, $B \in \mathbb{R}^{(n_2,k)}$ that solve:
\begin{equation}
    (A^*, B^*)  = \underset{A^T\Sigma_{1}A = B^T\Sigma_{2}B = I}{\arg\max} Tr (A^T \Sigma_{1,2}B)
\end{equation}
If we further have $k=n_1=n_2$ then the maximum value for $Tr (A^T \Sigma_{1,2}B)$ is $F(X_1,X_2) =Tr(T^{\top}T)^{\frac{1}{2}}$ with $T = \Sigma_1^{\frac{1}{2}}\Sigma_{1,2} \Sigma_{2}^{\frac{1}{2}}$. This value is the total CCA correlation of the random vectors $X_1,X_2$. It can also be seen as the sum of the singular values of $T$, each singular value representing the correlation of the embeddings along a direction. Note that this optimal value $F(X_1,X_2)$ only depends on the covariance matrices $(\Sigma_1, \Sigma_2, \Sigma_{1,2})$.

In the DCCA method, we consider two neural networks $g_1$, $g_2$ so as to optimize the total CCA correlation $F(g_1(X_1), g_2(X_2))$. The gradient of this objective with respect to the parameters of $g_1, g_2$ can be derived in order to use gradient descent.

In practice, to compute $F$ we can use the singular value decomposition of $T$ and sum the first $k$ singular values of $T$. Furthermore the singular values are interesting since they give an information of how much correlation is contained in each projection. That information can be used to analyse the data and choose an optimal dimension $k$ for the projection.

When considering more than two modalities, a proposed extension to the CCA is to optimize the sum of the pairwise CCA objectives \cite{kanatsoulis_mcca}. We adapt this idea to the DCCA framework and train DCCA encoders for $m$ modalities by maximizing $\sum_{i<j \in [|1,m|]} F(g_i(X_i),g_j(X_j))$.

Our implementation is based upon \url{https://github.com/Michaelvll/DeepCCA}.

\subsection{Multimodal Contrastive Learning}\label{contrastive learning}

Contrastive learning methods have emerged as a powerful tool to learn descriptive, transferable representations of high dimensional data such as images or text \cite{tian_constrastive_multiview_coding, clip_2021}. 

In the two-modalities case, we aim at learning two embbeding functions $g_1(x_1)$, $g_2(x_2)$ that brings together "positive pairs" observed from the joint distribution $x_1,x_2 \sim p(x_1,x_2)$ and separates "negatives pairs" observed from the product of the marginal distributions $x_1,x_2 \sim p(x_1)p(x_2)$. 

Formally, considering a batch of multimodal samples $(x_1^{i}, x_2^{i})_{1 \leq i \leq K}$, the loss function writes:

\begin{align}
    &L = \sum_{i=1}^{K} L_{1,2}(i) + L_{2,1}(i) \\
    &L_{1,2}(i) = - \log \left( \frac{sim_{\gamma}(x_1^{i},x_2^{i})}{\sum_{j = 1}^K sim_{\gamma}(x_1^{i},x_2^{j})} \right) \forall 1 \leq i \leq K \\
    &L_{2,1}(i) = - \log \left( \frac{sim_{\gamma}(x_2^{i},x_1^{i}))}{\sum_{j = 1}^K sim_{\gamma}(x_2^{i},x_1^{j})} \right) \forall 1 \leq i \leq K \,,
\end{align}
where $sim_{\gamma}(x_1,x_2) = \exp(\frac{1}{\tau} \frac{g_1(x_1)}{||g_1(x_1)||} \cdot \frac{g_2(x_2)}{||g_2(x_2)||})$ is the exponential cosine similarity between the embeddings,$\tau$ is a hyperparameter and $\gamma$ parameterize the embedding functions $g_1$, $g_2$ that we aim to optimize. $\tau = 0.1$ in our experiments.

For any $1 \leq i \leq K$, the pair $(x_1^{(i)}, x_2^{(i)})$ is a \emph{positive} pair which should have high similarity and the pairs $(x_1^{(i)}, x_2^{(j)})_{1\leq j\neq i \leq K}$, $(x_1^{(j)}, x_2^{(i)})_{1\leq j\neq i \leq K}$ are \emph{negative} pairs that should have low similarity.

In order to bring together positive pairs in the embedding space and separate negative pairs, the projectors $(g_j)_{1 \leq j \leq M}$ have to extract the information between modalities.

For a larger number of modalities: $M \geq 2$, we can compute the sum of all pairwise losses and minimize them jointly \cite{tian_constrastive_multiview_coding}. 

\section{Interpretations of the $\ljm$ Objective}
\label{interpretations}
In this appendix, we provide several interpretations of the $\ljm$ loss function Equation~\eqref{ljm}that explains why minimizing it is a sensible objective to fit the unimodal posteriors. First, we reinterpret Equation~\eqref{ljm} to show that it brings the unimodal encoder $q_{\phi_i}(z|x_i)$ (for $i\in[1,m]$) close to an average distribution $q_{\mathrm{avg}}(z|x_i) =  \mathbb{E}_{\hat{p}((x_j)_{j\neq i}|x_i)}(q_\phi(z|X))$ that is close to $ p_\theta(z|x_i)$ provided that the joint encoder is well fit. Secondly, we recall an analysis from \cite{suzuki_joint_2016} that links Equation~\eqref{ljm} to the notion of Variation of Information.

\subsection{Interpretation in Relation to an Average Distribution}
We recall an interpretation by \cite{vedantam_generative_2018} and extend it to a more general case.

First, let's suppose that we have only two modalities $x_1,x_2$ and that $x_2$ takes only discrete values in a set $V_2$. 
We isolate the term with $q_{\phi_2}(z|x_2)$ in Equation~\eqref{ljm} and sum over the whole dataset $D$:

\begin{align}
   \sum_{(x_1,x_2)\in D}^{N} KL(q_{\phi}(z|x_1,x_2) || q_{\phi_2}(z|x_2)) &= \sum_{y \in V_2} \sum_{(x_1,y) \in D} KL(q_{\phi}(z|x_1,y) || q_{\phi_2}(z|y)) \\
   &= \sum_{y \in V_2}  KL( \sum_{(x_1,y) \in D} q_{\phi}(z|x_1,y) || q_{\phi_2}(z|y))
\end{align}

Each distribution $ q_{\phi}(z|x_1,y)$ is a gaussian with a small variance, and $q_{\phi_2}(z|y)$ is encouraged to cover this mixture of all distributions $ \sum_{(x_1,y) \in D}q_{\phi}(z|x_1,y)$ which correspond to all parts of the latent space where a pair $(x_1,y)$ was embedded with the joint encoder $\jointpost$.

We now study the general case with $M \in \mathbb{N}$ and $x_j$ not taking discrete values. 

For $1 \leq j \leq M$, we isolate the term with $\qj$ in Equation~\eqref{ljm}: 
\begin{align}
     \ljm(j)  &= KL(\jointpost||\qj)\\
     &= \mathbb{E}_{\jointpost}\left( -\log(\qj)\right) - H(\jointpost)
\end{align}

where $H(\jointpost)$ is the Shannon entropy of $\jointpost$. 
Since, $\jointpost$ is fixed while optimizing Equation~x\eqref{ljm}, this term is a constant. 

When optimizing this loss over the entire dataset, we actually optimize the expectation of this term over the empirical distribution $p(X)$. 

\begin{align}
    \mathbb{E}_{p(X)}(\ljm(j)) = \mathbb{E}_{p(X)}\left( \mathbb{E}_{\jointpost}\left( -\log(\qj)\right)\right) + cte
\end{align}
where $cte$ is an additive constant term. 

Furthermore, we can decompose $p(X) = p(x_j)p(X_{C_j}|x_j)$ where we note $X_{C_j} = (x_i)_{1 \leq i \neq j \leq M}$ the set of modalities from which we exclude $x_j$. 

\begin{align}
    \mathbb{E}_{p(X)}(\ljm(j)) &= \mathbb{E}_{p(x_j)}\left( \mathbb{E}_{p(X_{C_j}|x_j)} \left( \mathbb{E}_{\jointpost}\left( -\log(\qj)\right)\right)\right) + cte
\end{align}

We suppose the density $\qj$ bounded by a constant $C$, which allows us to use Fubini's theorem and exchange the expectations.

\begin{align}
    \mathbb{E}_{p(X)}(\ljm(j)) &= \mathbb{E}_{p(x_j)}\left( \mathbb{E}_{p(X_{C_j}|x_j)} \left( \mathbb{E}_{\jointpost}\left( -\log(\frac{\qj}{C})\right)\right)\right) - \log(C) + cte\\
    &=\mathbb{E}_{p(x_j)}\left( \int_{X_{C_j}} \int_z -\log\left( \frac{\qj}{C}\right)
    \jointpost p(X_{C_j}|x_j)dz dX_{C_j} \right) +cte\\
    &= \mathbb{E}_{p(x_j)}\left( \int_z -\log\left( \frac{\qj}{C}\right)
    \int_{X_{C_j}} \jointpost p(X_{C_j}|x_j) dX_{C_j} dz \right) +cte\\ \label{fubini}
    &= \mathbb{E}_{p(x_j)}\left( \mathbb{E}_{q_{\phi}^{(avg)}(z|x_j)}\left(  -\log\left( \frac{\qj}{C}\right)\right) \right) +cte\\
    &= \mathbb{E}_{p(x_j)}\left( KL \left( q_{\phi}^{(avg)}(z|x_j) || \qj \right)\right) + H(q_{\phi}^{(avg)}(z|x_j)) +cte\\
\end{align}

where $q_{\phi}^{(avg)}(z|x_j) \coloneqq \int_{X_{C_j}} \jointpost p(X_{C_j}|x_j)dX_{C_j}$ and $cte$ regroups all additive constant terms at each line. We use Fubini's theorem at line \eqref{fubini} since all terms in the integral are positive.

Since $H(q_{\phi}^{(avg)}(z|x_j))$ is also a constant term, we see that minimizing $\ljm(j)$ reduces to miniming the Kullback-Leibler divergence between $\qj$ and this average distribution $q_{\phi}^{(avg)}(z|x_j)$.

\subsection{Interpretation in Relation to the Variation of Information}
First, in the bimodal case where $M=2$, we recall an interpretation provided by \cite{suzuki_joint_2016} that links \eqref{ljm} to the Variation of Information (VI) of $x_1$ and $x_2$ where $x_1$ (resp. $x_2$) represent the variable of the first modality (resp second).

Recall the definition of the VI : 
\begin{equation}
\label{vi def}
VI(x_1,x_2) = - \mathbb{E}_{\mathbb{P}(x_1,x_2)} \big(\log \mathbb{P}(x_1|x_2) + \log \mathbb{P}(x_2|x_1) \big)\,.
\end{equation}
If we analyse Eq.~\eqref{vi def}, we see that the more the modalities are predictive of one another, the smaller is the Variation of Information. 
We do not know the true joint and conditional distributions but we can use the following approximation summing on $N$ training samples:
\begin{equation*}
    \widetilde{VI} = - \sum_{n=1}^N \log p_{\theta,\phi_1}(x_1^{(n)}|x_2^{(n)}) + \log p_{\theta, \phi_2}(x_2^{(n)}|x_1^{(n)})\,,
\end{equation*}
where for $i,j \in \{1,2\}$ with $i\neq j$, $p_{\theta,\phi_i}(x_j|x_i) \vcentcolon= \int p_\theta(x_j|z) q_{\phi_i}(z|x_j)dz$ is our conditional generative models to sample $x_j$ from $x_i$. 
We can show that with $\mathcal{L}$ being the ELBO defined in Eq.~\eqref{elbo def} and $\ljm$ defined in Eq.~\eqref{ljm}:
\begin{equation}
\label{bound on vi}
    - \mathcal{L}(x_1,x_2;\theta,\phi) +\ljm(x_1,x_2;\phi) \geq \widetilde{VI}\,.
\end{equation}
We recall that in our method, we first maximise $\mathcal{L}(x_1,x_2; \theta, \phi)$ and then we minimize $\ljm(x_1,x_2;\phi)$, therefore we minimize an upper bound on $\Tilde{VI}$ that is the empirical Variation of Information between modality 1 and 2. Minimizing $\Tilde{VI}$ is a sensible goal as it encapsulates the predictive power of a modality given the other.

Let us now prove Equation~\eqref{bound on vi} :
\begin{small}
\begin{equation*}
    \begin{split}
        \log p_{\theta,\phi_1}(x_2|x_1) + \log p_{\theta,\phi_2}(x_1|x_2) &\geq \mathbb{E}_{q_\phi(z|x_1,x_2)} \left ( \log \frac{p_\theta(x_1|z)q_{\phi_2}(z|x_2)}{q_\phi(z|x_1,x_2)} \right ) \\ 
        &+ \mathbb{E}_{q_\phi(z|x_1,x_2)}\big( \log \frac{p_\theta(x_2|z)q_{\phi_1}(z|x_1)}{q_\phi(z|x_1,x_2)} \big)\\
        &= \mathbb{E}_{q_\phi(z|x_1,x_2)}\big(\log p_\theta(x_1|z)) + \mathbb{E}_{q_\phi(z|x_1,x_2)} \big(\log p_\theta(x_2|z) \big )\\ 
        &- KL(q_\phi(z|x_1,x_2) || q_{\phi_2}(z|x_2)) - KL(q_\phi(|x_1,x_2) || q_{\phi_1}(z|x_1))\\
        &= \mathcal{L}(x_1,x_2) + KL(q_\phi(z|x_1,x_2) ||p(z)) - \ljm(x_1,x_2; \theta, \phi)\\
        &\geq \mathcal{L}(x_1,x_2) - \ljm(x_1,x_2; \theta, \phi)\,.
    \end{split}
\end{equation*}
\end{small}

\section{Additional experimental results}\label{app:additional results}

\subsection{Additional results on MNIST-SVHN}

\begin{figure}[H]
    \centering
    \begin{subfigure}[b]{0.25\textwidth}
        \centering
        \includegraphics[width=\textwidth]{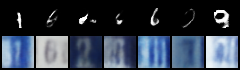}
        \caption{\tiny{JMVAE}}
    \end{subfigure}
    \hfill
    \begin{subfigure}[b]{0.25\textwidth}
        \centering
        \includegraphics[width=\textwidth]{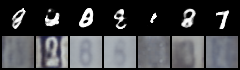}
        \caption{\tiny{MMVAE}}
    \end{subfigure}
    \hfill
    \begin{subfigure}[b]{0.25\textwidth}
        \centering
        \includegraphics[width=\textwidth]{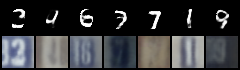}
        \caption{\tiny{MMVAE+}}
    \end{subfigure}
    \hfill
    \begin{subfigure}[b]{0.25\textwidth}
        \centering
        \includegraphics[width=\textwidth]{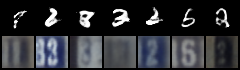}
        \caption{\tiny{MVTCAE}}
    \end{subfigure}
    \hfill
    \begin{subfigure}[b]{0.25\textwidth}
        \centering
        \includegraphics[width=\textwidth]{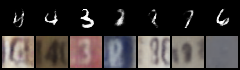}
        \caption{\tiny{MoPoE}}
    \end{subfigure}
    \hfill
    \begin{subfigure}[b]{0.25\textwidth}
        \centering
        \includegraphics[width=\textwidth]{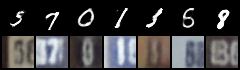}
        \caption{\tiny{JNF}}
    \end{subfigure}
    \hfill
    \begin{subfigure}[b]{0.25\textwidth}
        \centering
        \includegraphics[width=\textwidth]{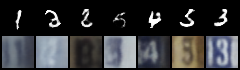}
        \caption{\tiny{JNF-CL}}
    \end{subfigure}
    \caption{Unconditional generation: for each model, latent codes are sampled from the prior and decoded jointly. }\label{fig:unconditional ms}
       
\end{figure}

In Figure~\ref{fig:unconditional ms}, we present samples generated from the prior. JNF-CL refers to our model JNF-Shared using Constrastive Learning (CL) to extract the shared information. This method performed best on this dataset, to extract the shared information.

In Table~\ref{tab:all results ms}, we report all coherences results for different values of the parameter $\beta$. For each model, we kept the value of $\beta$ that maximises the mean coherence for the results presented in Table~\ref{tab:ms_results}.

\begin{table}[H]
    \begin{centering}
    \scriptsize
\begin{tabular}{llrrrrrr}
    \hline
     &  & \multicolumn{2}{r}{Joint} & \multicolumn{2}{r}{$M \longrightarrow S$} & \multicolumn{2}{r}{$S \longrightarrow M$} \\
     &  & mean & std & mean & std & mean & std \\
    Model & $\beta$ &  &  &  &  &  &  \\
    \hline
    \multirow[t]{3}{*}{JMVAE} & 0.5 & 0.27 & 0.02 & 0.67 & 0.03 & 0.57 & 0.03 \\
     & 1 & 0.34 & 0.07 & 0.69 & 0.05 & 0.54 & 0.03 \\
     & \bf{2.5} & 0.43 & 0.10 & 0.73 & 0.07 & 0.53 & 0.05 \\
    \cline{1-8}
    \multirow[t]{3}{*}{MMVAE} & \bf{0.5} & 0.35 & 0.02 & 0.80 & 0.01 & 0.70 & 0.02 \\
     & 1 & 0.35 & 0.02 & 0.80 & 0.02 & 0.68 & 0.02 \\
     & 2.5 & 0.33 & 0.01 & 0.80 & 0.02 & 0.68 & 0.03 \\
    \cline{1-8}
    \multirow[t]{3}{*}{MMVAE+} & 0.5 & 0.24 & 0.04 & 0.55 & 0.04 & 0.62 & 0.02 \\
     & 1 & 0.27 & 0.03 & 0.50 & 0.03 & 0.59 & 0.06 \\
     & \bf{2.5} & 0.43 & 0.05 & 0.60 & 0.09 & 0.58 & 0.05 \\
    \cline{1-8}
    \multirow[t]{3}{*}{MVTCAE} & 0.5 & 0.29 & 0.01 & 0.74 & 0.02 & 0.36 & 0.02 \\
     & 1 & 0.35 & 0.02 & 0.75 & 0.05 & 0.44 & 0.02 \\
     & \bf{2.5 }& 0.44 & 0.02 & 0.81 & 0.01 & 0.52 & 0.02 \\
    \cline{1-8}
    \multirow[t]{3}{*}{MoPoE} & 0.5 & 0.27 & 0.02 & 0.13 & 0.01 & 0.77 & 0.00 \\
     & 1 & 0.32 & 0.01 & 0.12 & 0.00 & 0.75 & 0.01 \\
     & \bf{2.5} & 0.36 & 0.01 & 0.12 & 0.00 & 0.72 & 0.01 \\
    \cline{1-8}
    \multirow[t]{3}{*}{JNF} & 0.5 & 0.37 & 0.01 & 0.80 & 0.01 & 0.47 & 0.01 \\
     & 1 & 0.43 & 0.01 & 0.81 & 0.01 & 0.48 & 0.02 \\
     & \bf{2.5} & 0.51 & 0.01 & 0.82 & 0.01 & 0.52 & 0.01 \\
    \cline{1-8}
    \multirow[t]{3}{*}{JNF-Dcca} & 0.5 & 0.36 & 0.02 & 0.76 & 0.01 & 0.71 & 0.02 \\
     & 1 & 0.42 & 0.02 & 0.76 & 0.01 & 0.71 & 0.02 \\
     & \bf{2.5} & 0.51 & 0.01 & 0.75 & 0.03 & 0.69 & 0.05 \\
    \cline{1-8}
    \multirow[t]{3}{*}{JNF-CL} & 0.5 & 0.36 & 0.03 & 0.78 & 0.02 & 0.79 & 0.01 \\
     & 1 & 0.42 & 0.01 & 0.81 & 0.01 & 0.78 & 0.02 \\
     & \bf{2.5} & 0.51 & 0.02 & 0.81 & 0.01 & 0.75 & 0.02 \\
     \cline{1-8}
    \hline
    \end{tabular}
    \caption{All coherences results for different values of $\beta$ for each model. We indicate in bold, the value of $\beta$ that maximises average (conditional and joint) coherence for each model and that we kept for table \ref{tab:ms_results}. In \ref{tab:ms_results}, we presented results for our model JNF-Shared using Constrastive Learning (CL). Here we present additional results with the DCCA used instead of Constrastive Learning.}\label{tab:all results ms}
\end{centering}
\end{table}

\subsection{Additional results on PolyMNIST}
\label{mmnist:vis}

\begin{figure}[H]
    \centering
    \includegraphics[width = \textwidth]{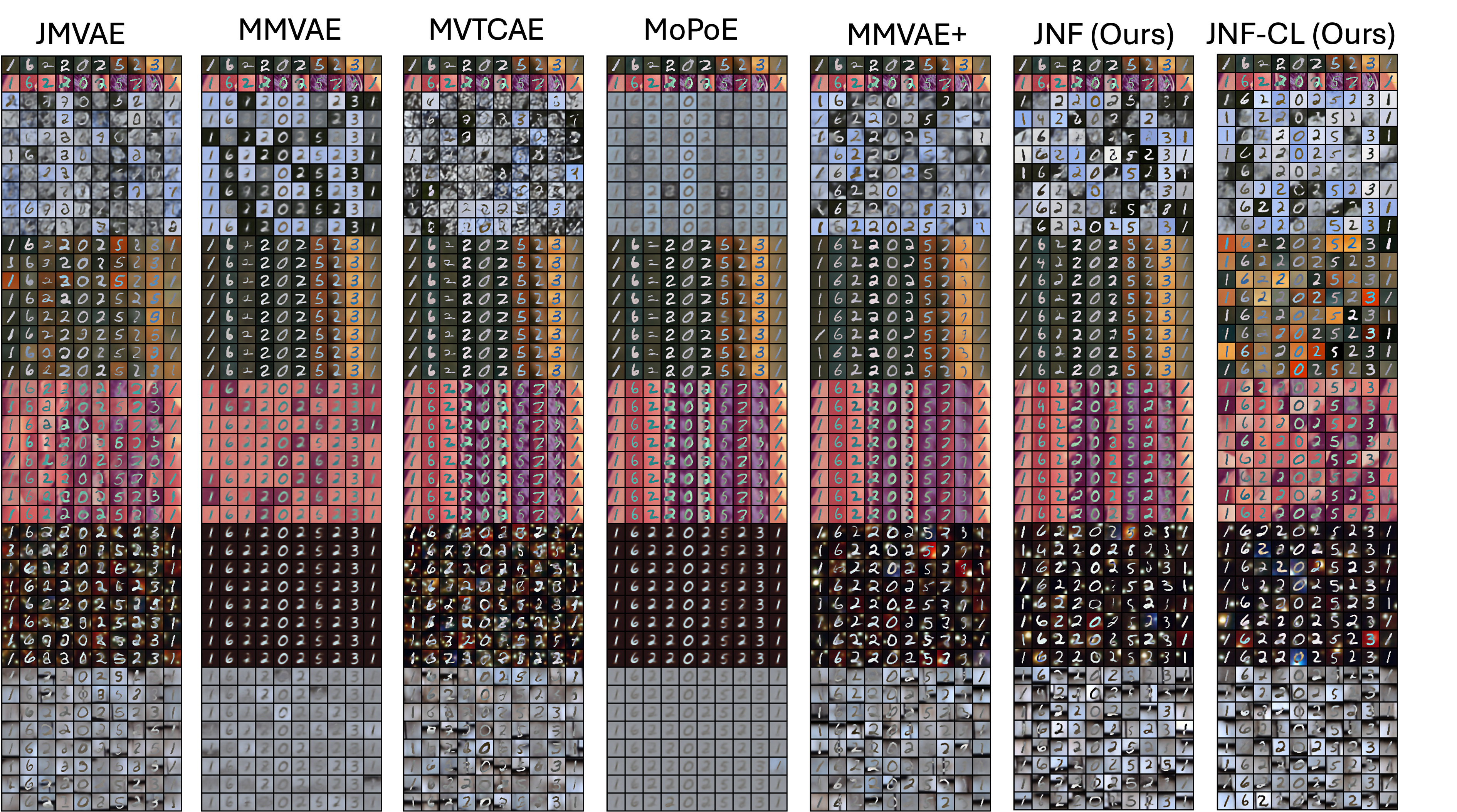}
    \caption{We present generated samples when conditioning on the first two modalities. 
    The first two rows are the samples we condition on and the rest of the rows are generated samples in each modality.}\label{fig: mmnist cond samples}
\end{figure}

In Figure~\ref{fig: mmnist cond samples} we present samples generated by conditioning on a subset of two modalities and in Figure~\ref{fig:mmnist unconditional} we present samples generated from the prior (unconditional generation). Our models produce diverse and coherent images, while the MoPoE and MMVAE models produce images that look "averaged" from using a mixture based aggregation \cite{daunhawer_limitations_2022}. 

\subsection{Additional results on Translated PolyMNIST}
\begin{figure}[H]
    \centering
    \includegraphics*[width=\textwidth]{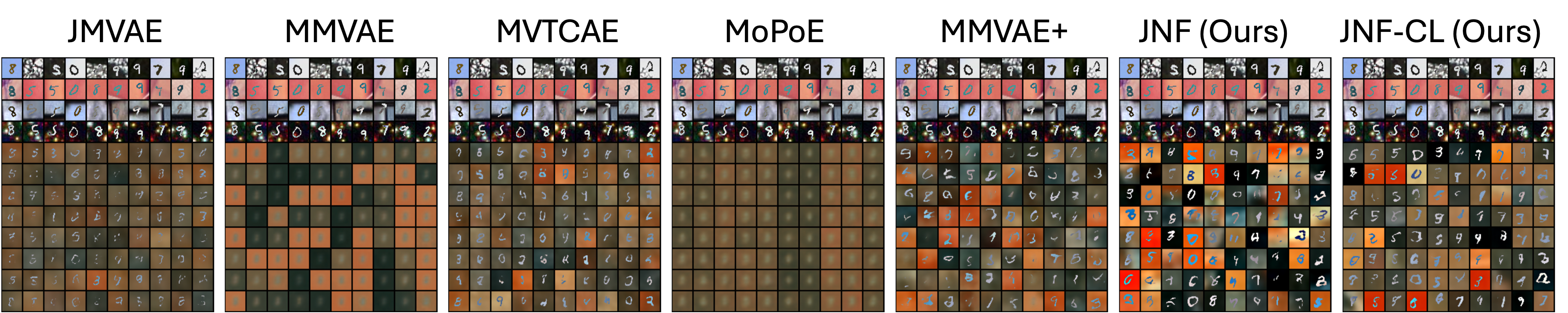}
    \caption{Conditional generation on Translated PolyMNIST. The first four rows are the images we condition on and the newt rows are generated samples in the first modality. JNF-CL refers to our model JNF-Shared with CL.}\label{fig:conditional generation on tmmnist}
\end{figure}

\begin{figure}[H]
    \centering
    \begin{subfigure}[b]{0.19\textwidth}
        \centering
        \includegraphics[width=\textwidth]{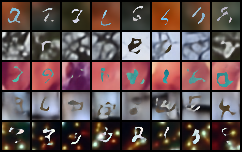}
        \caption{\scriptsize{JMVAE}}
    \end{subfigure}
    \begin{subfigure}[b]{0.19\textwidth}
        \centering
        \includegraphics[width=\textwidth]{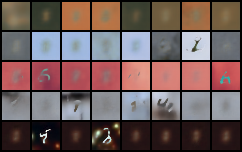}
        \caption{\scriptsize{MMVAE}}
    \end{subfigure}
    \begin{subfigure}[b]{0.19\textwidth}
        \centering
        \includegraphics[width=\textwidth]{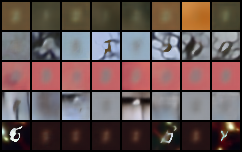}
        \caption{\scriptsize{MoPoE}}
    \end{subfigure}
    \begin{subfigure}[b]{0.19\textwidth}
        \centering
        \includegraphics[width=\textwidth]{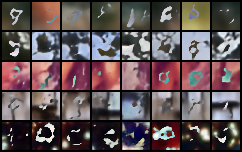}
        \caption{\scriptsize{MVTCAE}}
    \end{subfigure}
    \begin{subfigure}[b]{0.19\textwidth}
        \centering
        \includegraphics[width=\textwidth]{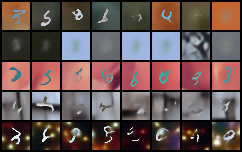}
        \caption{\scriptsize{MMVAE+}}
    \end{subfigure}
    \begin{subfigure}[b]{0.19\textwidth}
        \centering
        \includegraphics[width=\textwidth]{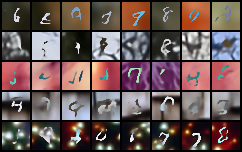}
        \caption{\scriptsize{JNF(Ours)}}
    \end{subfigure}
    \begin{subfigure}[b]{0.19\textwidth}
        \centering
        \includegraphics[width=\textwidth]{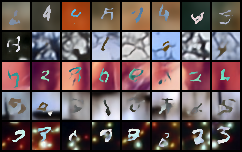}
        \caption{\scriptsize{JNF-CL (Ours)}}
    \end{subfigure}
    \caption{Unconditional generation on Translated PolyMNIST when sampling a latent code from the prior.}\label{fig:tmmnist unconditional app}
\end{figure}

Figure~\ref{fig:conditional generation on tmmnist} shows examples of generated images on TranslatedPolyMNIST and in Figure~\ref{fig:tmmnist unconditional app} we present samples generated from the prior. 
MMVAE and MoPoE reach a high joint coherence on this dataset but if we look at the generated images, we realize the generated images all look averaged, resembling a small "1" digit. The FID is very high since the generation is not diverse.

In Table~\ref{tab:all results tmmnist}, we present coherences and FID results for different values of the parameter $\beta$ for each model. We used this table for selecting the value of $\beta$. For all models we observe inverse tendencies between joint and conditional coherence with the value of $\beta$. We chose to favor conditional coherence to select the best value of $\beta$ for each model for the results presented in Table~\ref{fig:mmnist results}. In this table, we also test two values for the number of flows $n_{flows} \in \{ 2,3 \}$ for our models. \emph{After} selecting $\beta$, we varied and selected the optimal parameter $n_{flows}$. 

\begin{table}[H]
    \centering
    \footnotesize
    \begin{tabular}{lllrrr}
        \toprule
        & & & \multicolumn{2}{c}{Coherence $(\uparrow)$} & FID $(\downarrow)$ \\
         & & & Joint  &  Conditional & 1 modality to $m_0$ \\
        Model & $\beta$ & $n_{flows}$&  &  &  \\
        \cline{1-6}
        \multirow[t]{3}{*}{JMVAE}& \bf{0.5} & & 0.00 & 0.15 & 37.06 \\
         & 1.0 & & 0.00 & 0.14 & 43.93 \\
        &  2.5 & & 0.00 & 0.12 & 55.09 \\
        \cline{1-6}
        \multirow[t]{3}{*}{MMVAE+} & \bf{0.5}& & 0.006 & 0.10 & 60.48 \\
         & 1 & & 0.005 & 0.10 & 69.80 \\
         & 2.5& & 0.15 & 0.10 & 206.13 \\
        \cline{1-6}
        \multirow[t]{3}{*}{MVTCAE} & \bf{0.5}& & 0.004 & 0.13 & 42.35 \\
         & 1& & 0.08 & 0.11 & 121.86 \\
         & 2.5& & 0.23 & 0.11 & 178.49 \\
         \cline{1-6}
         \multirow[t]{3}{*}{MMVAE} & \bf{0.5}& & 0.63 & 0.10 & 185.97 \\
        & 1.0& & 0.49 & 0.10 & 172.44 \\
        & 2.5& & 0.58 & 0.10 & 181.08 \\
        \cline{1-6}
        \multirow[t]{3}{*}{MoPoE} & 0.5& & 0.26 & 0.10 & 195.53 \\
        & \bf{1.0}& & 0.50 & 0.10 & 199.48 \\
        & 2.5& & 0.50 & 0.10 & 199.94 \\
        \cline{1-6}
        \multirow[t]{3}{*}{JNF (Ours)} & \multirow[t]{2}{*}{\bf{0.5}} & 3  & 0.0004 & 0.17 & 30.91 \\
        &\bf{0.5} & \bf{2} & 0.0002 & 0.18& 31.76 \\
         & 1.0 & 3 & 0.0007 & 0.17 & 33.82 \\
         & 2.5 & 3 & 0.06 & 0.12 & 218.75 \\
        \cline{1-6}
        \multirow[t]{3}{*}{JNF-Shared (CL) (Ours)} & \bf{0.5}& 3 & 0.0002 & 0.21 & 32.09 \\
        &\bf{0.5} & \bf{2} & 0.0005 & 0.23 & 33.17\\
         & 1 & 3 & 0.0008 & 0.20 & 35.30 \\
         & 2.5& 3 & 0.06 & 0.13 & 217.02 \\
        \cline{1-6}
        \bottomrule
        \end{tabular}        
    \caption{Coherences and FID results for different values of $\beta$. Here, we average over all possible subsets for the conditional coherence. For almost all models, we observe inverse tendencies for joint and conditional coherence with the value of $\beta$. For the results presented in the main text, we chose to favor conditional generation to select the value of $\beta$ for each model. The chosen $\beta$ is set in bold. For the JNF-Shared (DCCA) we used the same $\beta = 0.5$ and 2 flows as for JNF-Shared (CL) because of the similarity between models.    }\label{tab:all results tmmnist}
\end{table}        

\subsection{Additional results on the MHD dataset}
\label{mhd:vis}

\begin{figure}[htb]
    \centering
    \includegraphics[width=\textwidth]{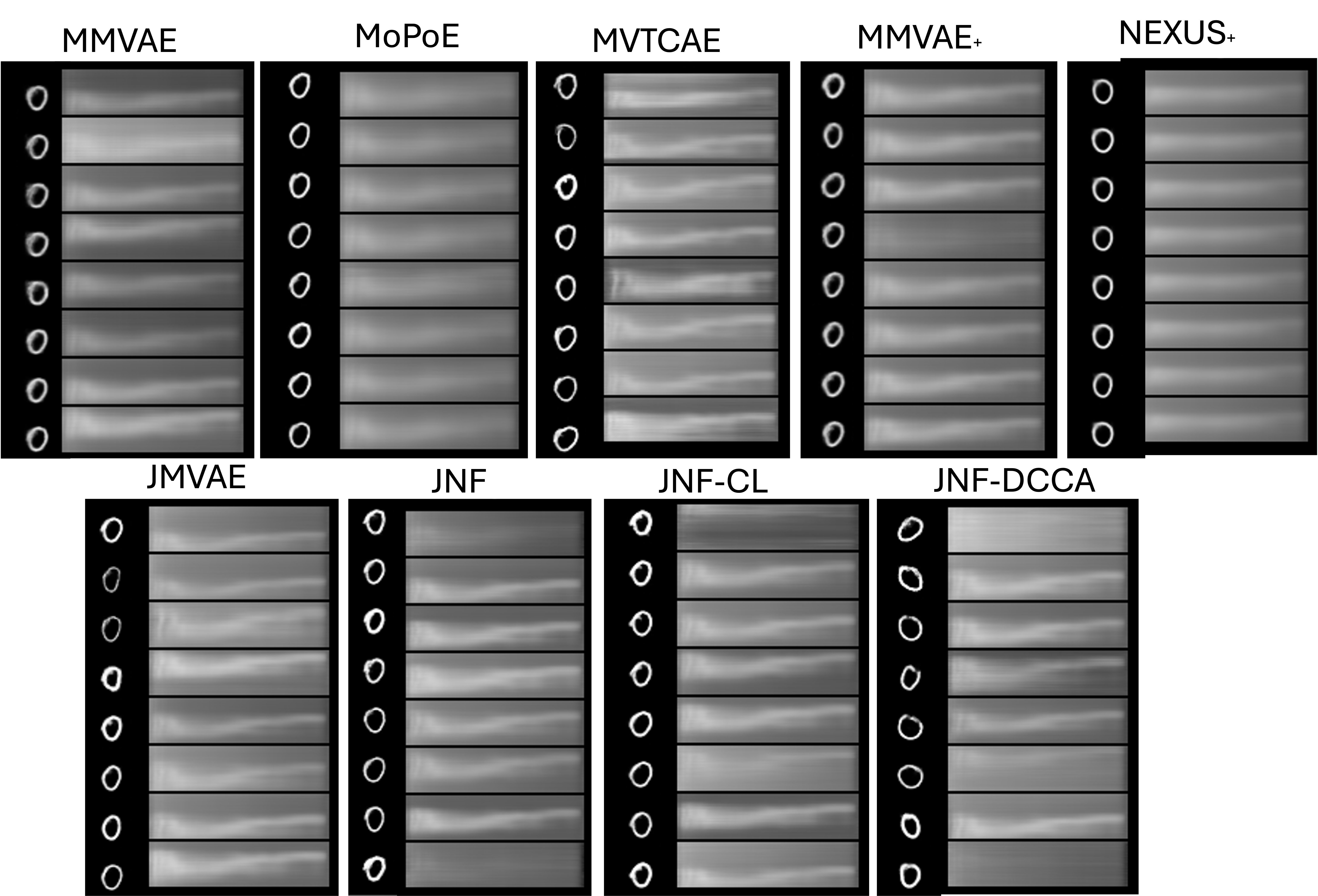}
    \caption{Samples generated when conditioning on a given trajectory.}\label{fig:conditional samples mhd}
\end{figure}

In Figure~\ref{fig:conditional samples mhd}, we display images and spectrograms obtained when conditioning on a given trajectory (that is not displayed here) drawing a zero digit. Our models generate diverse and constrasted images.

In Table~\ref{tab:all results mhd}, we present all coherence results for different values of the parameter $\beta$ for each model. We used this table to chose $\beta$ for each model. For all models we observe inverse tendencies between joint and conditional coherence with the value of $\beta$. We chose to favor conditional coherence to select $\beta$ for each model for the results presented in Table~\ref{fig:mhd_results}.

\begin{table}
    \centering
    \footnotesize
\begin{tabular}{llrrrr}
    \toprule
     &  & \multicolumn{2}{r}{Joint} & \multicolumn{2}{r}{Conditional} \\
     &  & mean & std & mean & std \\
    Model & $\beta$ &  &  &  &  \\
    \midrule
    \multirow[t]{3}{*}{JMVAE} & \bf{0.5} & 0.57 & 0.02 & 0.86 & 0.01 \\
     & 1.0 & 0.15 & 0.02 & 0.79 & 0.01 \\
     & 2.5 & 0.15 & 0.04 & 0.75 & 0.02 \\
    \cline{1-6}
    \multirow[t]{3}{*}{MMVAE} & \bf{0.5} & 0.63 & 0.01 & 0.86 & 0.01  \\
    & 1.0 & 0.60 & 0.07 & 0.84 & 0.04  \\
    & 2.5 & 0.65 & 0.02 & 0.86 & 0.01 \\
    \cline{1-6}
    \multirow[t]{3}{*}{MMVAEPlus} & \bf{0.5} & 0.58 & 0.03 & 0.89 & 0.02 \\
     & 1.0 & 0.64 & 0.05 & 0.82 & 0.03 \\
     & 2.5 & 0.47 & 0.15 & 0.50 & 0.08 \\
    \cline{1-6}
    \multirow[t]{3}{*}{MVTCAE} & \bf{0.5} & 0.38 & 0.01 & 0.87 & 0.01 \\
     & 1.0 & 0.48 & 0.01 & 0.85 & 0.00 \\
     & 2.5 & 0.54 & 0.02 & 0.79 & 0.01 \\
    \cline{1-6}
    \multirow[t]{3}{*}{MoPoE} & \bf{0.5} & 0.44 & 0.02 & 0.74 & 0.01 \\
     & 1.0 & 0.50 & 0.01 & 0.72 & 0.02 \\
     & 2.5 & 0.45 & 0.02 & 0.62 & 0.01 \\
    \cline{1-6}
    \multirow[t]{3}{*}{JNF} & \bf{0.5} & 0.67 & 0.01 & 0.89 & 0.01 \\
     & 1.0 & 0.71 & 0.02 & 0.86 & 0.01 \\
     & 2.5 & 0.72 & 0.02 & 0.81 & 0.01 \\
    \cline{1-6}
    \multirow[t]{3}{*}{JNF-Shared (DCCA)} & \bf{0.5} & 0.66 & 0.01 & 0.92 & 0.01 \\
     & 1.0 & 0.71 & 0.02 & 0.90 & 0.01 \\
     & 2.5 & 0.72 & 0.02 & 0.80 & 0.01 \\
     \cline{1-6}
     \multirow[t]{1}{*}{JNF-Shared (CL)} & \bf{0.5} & 0.65 & 0.02 & 0.93 & 0.01 \\
    \cline{1-6}
    \bottomrule
    \end{tabular}
    \caption{All coherence results for different values of $\beta$ for each model. We indicate in bold, the value of $\beta$ that maximises conditional coherence for each model and that we kept for Table~\ref{fig:mhd_results}. For the JNF-Shared (CL) we use the same $\beta$ as for JNF-Shared (DCCA) since the model are very similar.}\label{tab:all results mhd}.
\end{table}
\newpage

\section{Architectures and hyperparameters used in the experiments}
\label{app:exp}

\begin{figure}[htb] 
    \centering 
    \begin{minipage}[b]{0.4\linewidth}
      \centering
      \includegraphics[width=\linewidth]{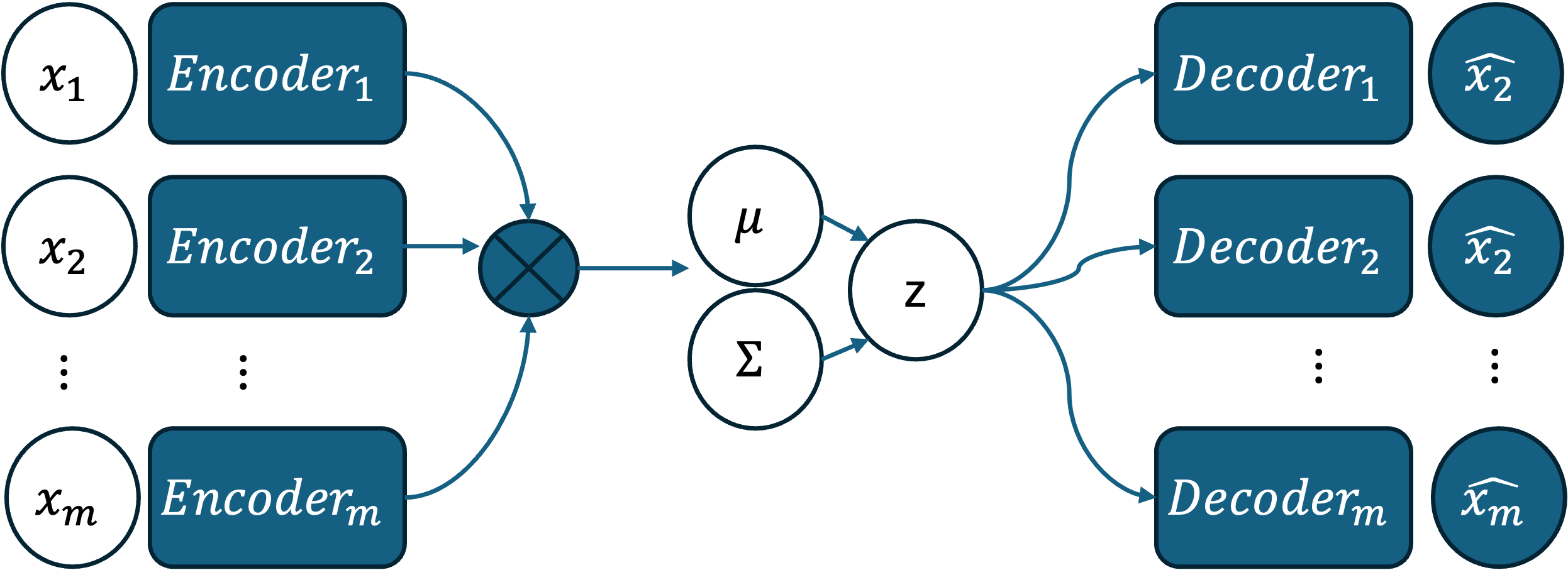} 
      \caption{Architectures of MMVAE, MoPoE and MVTCAE} 
    \end{minipage}
    \vspace{1cm}
    \begin{minipage}[b]{0.4\linewidth}
      \centering
      \includegraphics[width=\linewidth]{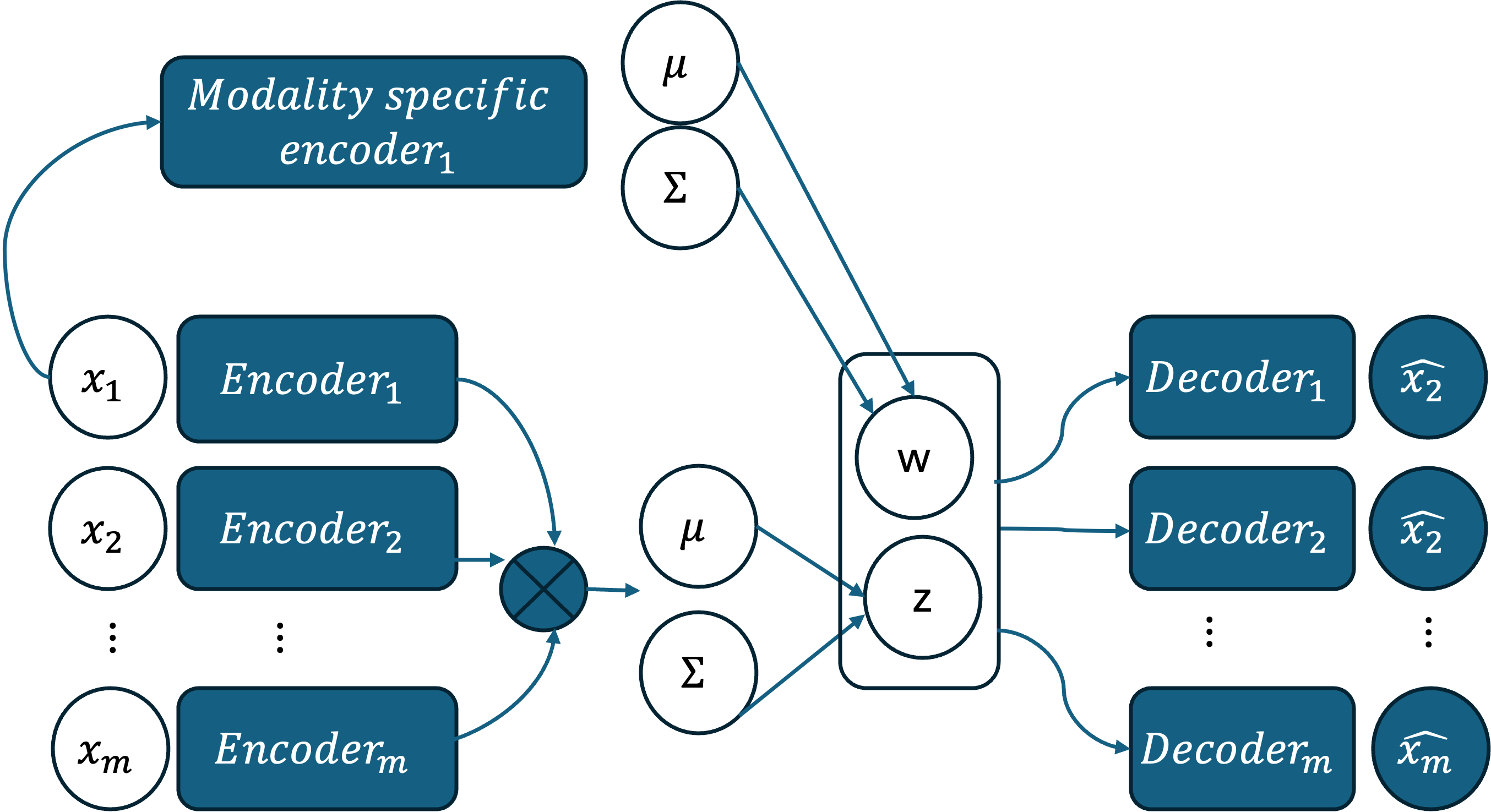} 
      \caption{Architecture of MMVAE+} 
      
    \end{minipage} 

    \begin{minipage}[b]{0.4\linewidth}
      \centering
      \includegraphics[width=\linewidth]{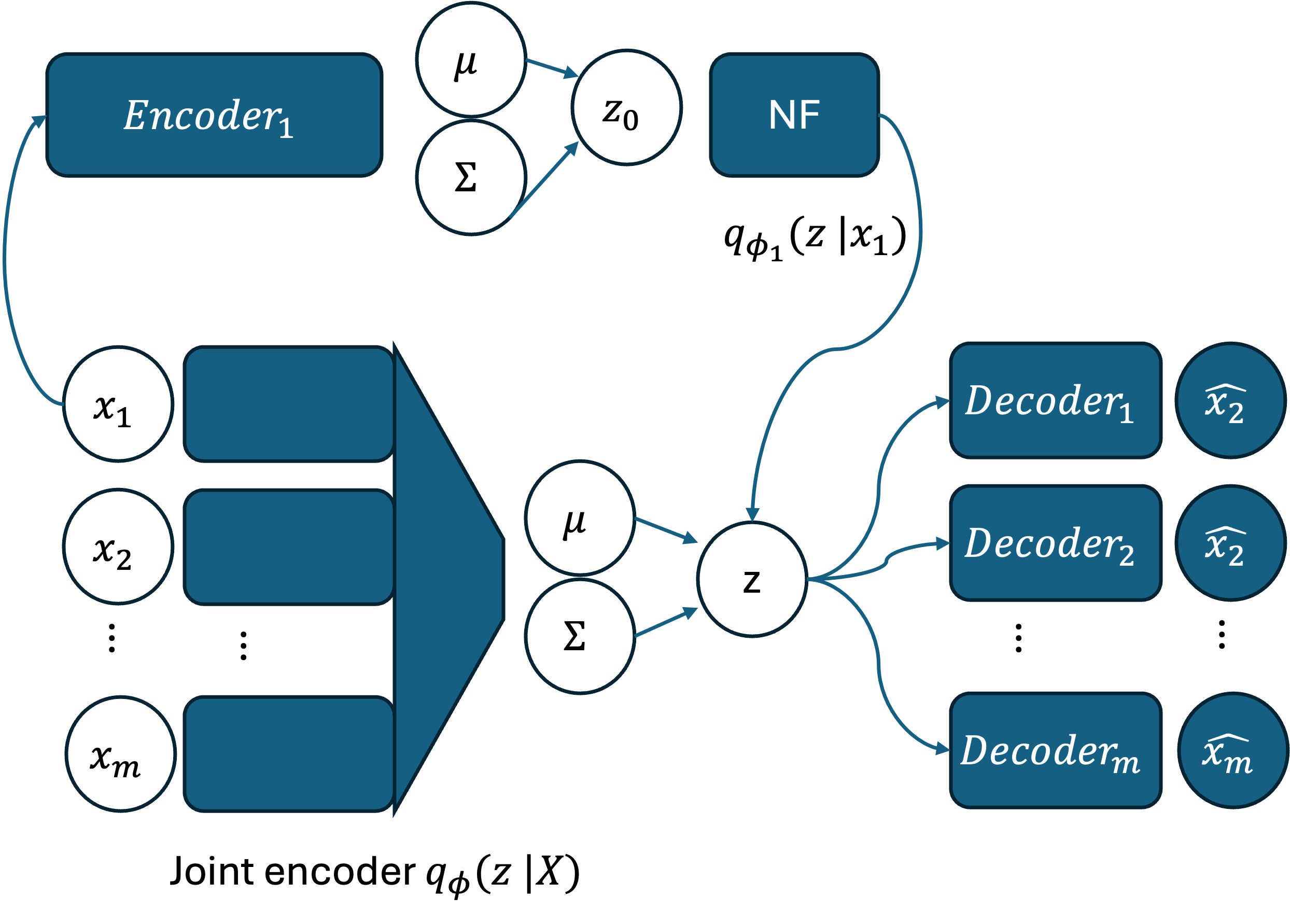} 
      \caption{Architecture of JNF and JMVAE model.} 
      
    \end{minipage}
    \vspace{1cm}
    \begin{minipage}[b]{0.4\linewidth}
      \centering
      \includegraphics[width=\linewidth]{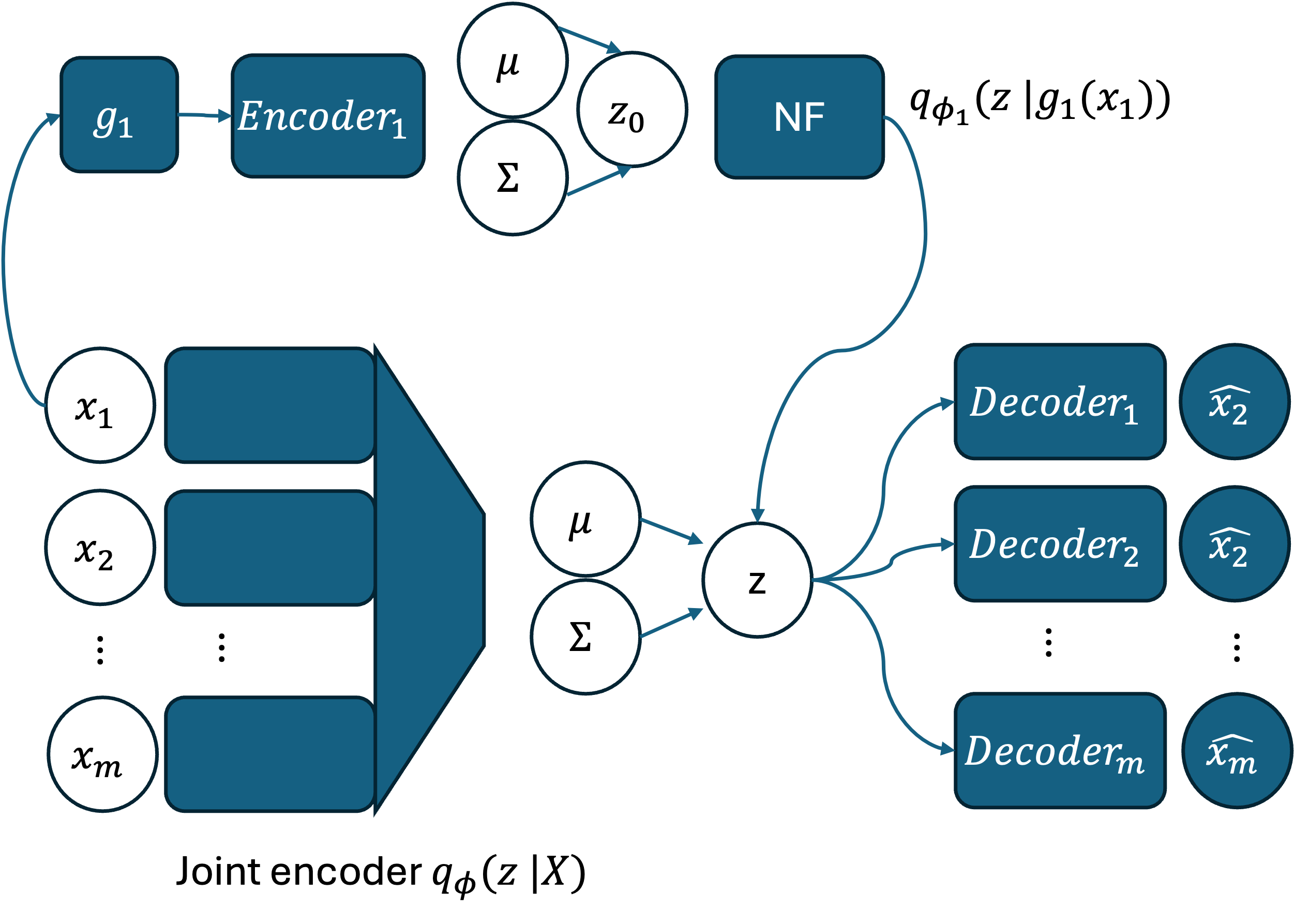} 
      \caption{Architecture of JNF-Shared} 
    \end{minipage} 
    \caption{Architectures for most models used. For the JMVAE model, it is the same architecture as the JNF model except without the Normalizing Flows.}\label{fig:archi}
  \end{figure}

In Figure~\ref{fig:archi} we summarize the general architectures of most models used in our experiments. 
For the Nexus model, we refer the reader to~\cite{vasco_leveraging_2022}.

We describe in the following sections the encoders/decoders architectures for all experiments. Note that for the JNF-Shared, the projectors $(g_j)_{1 \leq j \leq M}$ have the same architectures as the encoders of other models, and the encoders that parameterize $\qgj$ are simple two-layers MLPs taking the projections $(g_j)_{1 \leq j \leq M}(x_j)$ as inputs.

Our implementations of Normalizing Flows rely on the opensource library Pythae~\cite{chadebec_pythae_2022}. 

Code and data needed for reproducing the experiments are available at \url{https://anonymous.4open.science/r/JNF_VAE/README.md}.

\subsection{On MNIST-SVHN}

In Table~\ref{tab:parameters ms}, we indicate all architectures and training parameters used in the MNIST-SVHN experiments.
All models are trained until convergence. For all models, we test three values for $\beta \in \{0.5,1.0,2.5\}$ and for each model we kept the value that maximized average coherence (joint and conditional). Extensive results for all values of $\beta$ are presented in Table~\ref{tab:all results ms}. For the MMVAE and MMVAE+ model, we use Laplace distributions for modeling prior and posterior distribution following \cite{shi_variational_2019}. For all others models, we use Gaussian distributions for prior and posteriors.
For the decoders distributions $p_{\theta}(X|z)$ we use Laplace distributions. Following previous work \cite{shi_variational_2019,sutter_generalized_2021} we rescale the likelihoods of each modality with factors $\lambda_{MNIST}, \lambda_{SVHN}$ in order to compensate for the different sizes of the modalities and mitigate conflictual gradients \cite{javaloy_mitigating_2022}.The values for $\lambda_{MNIST}, \lambda_{SVHN}$ are indicated in Table~\ref{tab:parameters ms}. Intuitively, we need to put more weight on the smaller modalities so that they are also well reconstructed. 

We give specific details for each model:
\begin{itemize}

\item MVTCAE: we set $\alpha=0.9$ following their recommandations in the supplemental material in \cite{hwang_multi-view_2021}. 

\item MMVAE: we set K=10 for the number of samples in the ELBO.
\item MMVAE+: we set K=10 for the number of samples in the ELBO. The shared latent space as well as the modality-specific latent spaces have a dimension of 10.
\item JMVAE model, we set $\alpha=0.1$ as it appears as a good compromise value in \cite{suzuki_joint_2016}. We use annealing as in the original paper with a 100 epochs for warmup. The joint encoder is made up of separate heads and a common merging part where the separate heads have the same architecture as the unimodal encoders in Table \ref{tab:parameters ms}. The merging part is a simple two-layer MLP with 512 neurons in each layer.
\item JNF: we used Masked Autoregressive Flows with two MADE blocks\cite{papamakarios_masked_2017}. We use the same joint encoder as for the JMVAE model.
\item JNF-Shared: We use the same flows and joint encoder as JNF. The projectors used for CL or DCCA have the same architectures as the encoders in Table~\ref{tab:parameters ms}. The encoders $q_{\phi_j}(z|g_j(x_j))$ are simple networks with two linear hidden layers.
\end{itemize}

\begin{sidewaystable}[p]
    \footnotesize
    \begin{tabular}{@{}ll@{}}
        \toprule
        \headcol Mnist Encoder & Mnist Decoder \\
        \midrule
        Linear($1 \times 28 \times 28$ ,400) RELU & Linear(20, 400) RELU\\
        Linear(400,20), Linear(400,20) &Linear(400, $1 \times 28 \times 28$), Sigmoid\\
        
        \midrule
        \headcol  SVHN Encoder & SVHN Decoder \\
        \hline
        Conv2d(3,32,4,2,1,bias=True), RELU &  Conv2dTranspose(d,128,4,4,1,0,bias=True), RELU\\
        Conv2d(32,64,4,2,1,bias=True), RELU& Conv2dTranspose(128,64,4,4,1,0,bias=True), RELU\\
        Conv2d(64,128,4,2,1,bias=True), RELU& Conv2dTranspose(64,32,4,4,1,0,bias=True), RELU\\
        Conv2d(128,20,4,1,0,bias=True) $\times 2$ & Conv2dTranspose(32,3,4,4,1,0,bias=True), Sigmoid\\
        
        \headcol  Training parameters &  Joint Encoder for JMVAE, JNF, JNF-Shared \\
    \midrule
    Batchsize = 128 & Separates head for each modality\\
    Learning rate = 1e-3 & Linear(512), RELU\\
    Optimizer = Adam & Linear(512), RELU \\
    Latent dimension = 20& Linear(20), Linear(20) \\
    $\lambda_{MNIST}, \lambda_{SVHN} = \frac{3\times32\times32}{28\times28},1$ & \\
    Dimension for the projection $(g_j)$ = 10 &\\
    \midrule
    \headcol  Normaling Flows & JNF-Shared encoders for $\qgj$\\
    \midrule
    Masked Autoregressive with two MADE blocks  & Linear(10,512) RELU\\
                                                & Linear(512,512) RELU \\
                                                & Linear(512,20), Linear(512,20)\\
    \bottomrule
    
\end{tabular}
\caption{Architectures and training parameters used for the Mnist-SVHN Experiments. }\label{tab:parameters ms}
\end{sidewaystable}

\subsection{On PolyMNIST}

\begin{figure}[H]
    \includegraphics*[width=\textwidth]{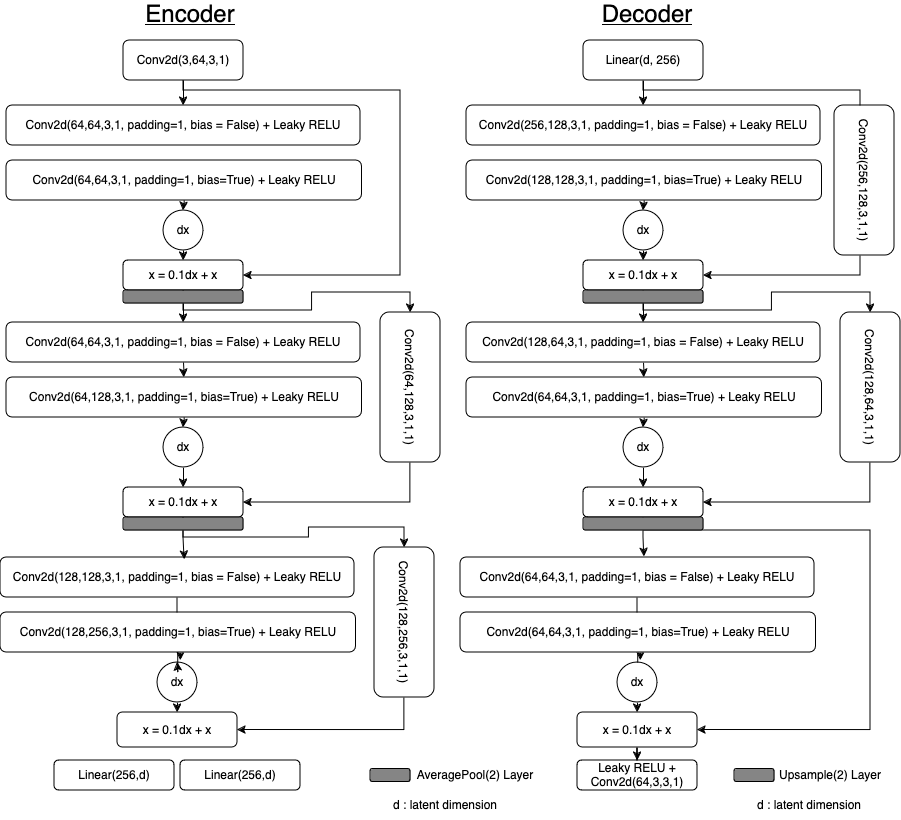}
    \caption{Encoder and decoder architectures used for the experiments on the PolyMNIST dataset.}\label{fig:resnets}
\end{figure}

For the PolyMNIST experiments, we used the same Resnet \cite{He2015DeepRL} architectures as used in \cite{palumbo_mmvae_2023}. These architectures are summarized in Figure~\ref{fig:resnets}.
Following \cite{palumbo_mmvae_2023}, we train all models as $\beta$-VAE and set $\beta = 2.5$. Each model is trained until convergence with a batchsize of 128 and learning rate of 1e-3. The latent dimension is set to 190 to match the total capacity of the MMVAE+ model in~\cite{palumbo_mmvae_2023}.

We give specific details for each model:
\begin{itemize}
    \item MMVAE: Due to memory limitations, we set the latent dim to 64 and used K=10 for the number of samples in the ELBO.
    \item MVTCAE: we set $\alpha = \frac{5}{6}$.
    \item JMVAE: we set $\alpha=0.1$ and annealing with a warmup of 100 epochs. In the original JMVAE model, a new encoder network needs to be introduced for each subset of modalities. In our experiments, we didn't choose that solution since it represents a very large number of parameters. Instead, we use for the JMVAE model, the PoE sampling solution that we also use for our models (Equation~\eqref{eq:poe}). The joint encoder is made-up of separate heads with the same architectures as in Figure~\ref{fig:resnets} and a merging neural networks with two hidden linear layers of 512 neurons. 
    \item MMVAE+: We use 32 dimensions for the shared latent space and 32 dimension for each modality specific space as in \cite{palumbo_mmvae_2023}.
    \item JNF: Same joint encoder as JMVAE. We use Masked Autoregressive flows with 2 MADE blocks.
    \item JNF-Shared: Same joint encoder and Normalizing flows as JNF. The projectors $(g_j)$ are simple convolutional networks similar to the SVHN encoders in~\ref{tab:parameters ms} and the encoders $\qgj$ are simple linear encoders as for the MNIST-SVHN experiments: see Table~\ref{tab:parameters ms}. 
\end{itemize}

\subsection{On Translated PolyMNIST}

For the TranslatedPolyMNIST experiments, we used similar architectures as in the PolyMNIST experiments with a latent dimension of 200 (except for MMVAE and MMVAE+ whose parameters are specified below).
We performed experiments with $\beta \in \{0.5, 1., 2.5\}$. For all models, we kept the value of $\beta$ that maximized average conditional coherence. In Table~\ref{tab:all results tmmnist}, we present results for different values of $\beta$ and the selected values for each model.  We use a latent dimension of 200 for all models but the MMVAE+ that has multiple latent spaces (see below). All models are trained until convergence with learning rate 1e-3 and batchsize 128.

We give specific details for each model:
\begin{itemize}
    \item MMVAE: Due to memory limitations, we used a latent dimension of 100 for the MMVAE model and used K=10 for the number of samples in the ELBO.
    \item MVTCAE: we set $\alpha = \frac{5}{6}$ as in PolyMNIST.
    \item JMVAE: we set $\alpha=0.1$ and a warmup of 100 epochs. PoE sampling is applied for JMVAE as in the other experiments. The joint encoder is made of separate heads with the same architectures as in Figure~\ref{fig:resnets} and we concatenate the outputs of each head to form the joint representation. This concatenation instead of a merging network allows to avoid conflictual gradient issues and modality collapse~\cite{javaloy_mitigating_2022}.  
    \item MMVAE+: We use 32 dimensions for the shared latent space and 32 dimension for each modality specific space as in \cite{palumbo_mmvae_2023}. We use K=10 for the number of samples in the ELBO.
    \item JNF: Same joint encoder as JMVAE. We use Masked Autoregressive flows with 3 MADE blocks.
    \item JNF-Shared (CL): Same joint encoder and Normalizing flows as JNF. The projectors $(g_j)$ have the encoder architectures in Figure~\ref{fig:resnets} and the encoders $\qgj$ are simple linear networks as for the MNIST-SVHN experiments: see Table~\ref{tab:parameters ms}. We use Masked Autoregressive flows with 2 MADE blocks.
    \item JNF-Shared (DCCA): when using DCCA to extract the shared information, we used more simple architectures for the projectors $(g_j)$ for instability reasons. We used simple convolutional networks similar to the SVHN encoders in Table~\ref{tab:parameters ms}. Precise architectures are given in the code. We use Masked Autoregressive flows with 2 MADE blocks.
\end{itemize}

\subsection{On MHD}
Table~\ref{tab:mhd parameters} contains all relevant architectures and general training parameters.  

We use the same architectures than the ones used in \cite{vasco_leveraging_2022} except that we don't pretrain the sound encoder and decoder. All models with a $\beta$ term weighing the Kullback-Leibler divergence in \eqref{elbo def} and for all models, the $\beta = 0.5$ gives the best average conditional coherence. We present additional results for all values of $\beta$ in Table~\ref{tab:all results mhd}.
We used Gaussian distributions to model all posterior, prior and decoding distributions. 
We use a latent dimension of 64 for all models but the MMVAE+ that has multiple latent spaces (see below).

We use rescaling for the likelihoods of each modality following \cite{shi_variational_2019}. It has been shown that this limits the phenomenons of conflictual gradients and modality collapse \cite{javaloy_mitigating_2022}. The rescaling factors $\lambda_{image}, \lambda_{audio},\lambda_{trajectory}$ are given in Table \ref{tab:mhd parameters}

We train all models until convergence.
We give specific details for each model:
\begin{itemize}
    \item MMVAE: We used K=10 for the number of samples in the ELBO.
    \item MVTCAE: we tried $\alpha \in \{0.75,0.9\}$ and kept best results obtained for $\alpha = 0.9$.
    \item JMVAE: we set $\alpha=0.1$ and a warmup of 100 epochs. In the original JMVAE model, a new encoder network needs to be introduced for each subset of modalities. In our experiments, we didn't choose that solution since it represents a very large number of parameters. Instead, we use for the JMVAE model, the PoE sampling solution that we also use for our models (Equation~\eqref{eq:poe}). The joint encoder is made-up of separate heads with the same architectures as in Table \ref{tab:mhd parameters} and a merging neural networks with two hidden linear layers of 512 neurons. 
    \item MMVAE+: We use 32 dimensions for the shared latent space and 32 dimension for each modality specific space. We used K=10 for the number of samples in the ELBO.
    \item JNF: Same joint encoder as JMVAE. We use Masked Autoregressive flows with 2 MADE blocks.
    \item JNF-Shared: Same joint encoder and Normalizing flows as JNF. The projectors $(g_j)$ have the encoder architectures in Figure~\ref{fig:resnets} and the encoders $\qgj$ have the same architectures as for the MNIST-SVHN experiments: see Table~\ref{tab:parameters ms}. 
    \item NEXUS : we use the same hyperparameters as used in \cite{vasco_leveraging_2022}.

\end{itemize}

\begin{sidewaystable}[hp]
    \footnotesize
    \begin{tabular}{@{}ll@{}}
        \toprule
        \headcol Image Encoder & Image Decoder \\
        \midrule
        Conv2d(1,128,4,2), Batchnorm, RELU & Linear(1024*4*4)\\
        Conv2d(128,256,4,2), Batchnorm, RELU &ConvTranspose2d(1024,512,3,2,padding = 1), Batchnorm,RELU\\
        Conv2d(256,512,4,2), Batchnorm, RELU & ConvTranspose2d(512,256,3,2,padding = 1),Batchnorm, RELU\\
        Conv2d(512,1024,4,2), Batchnorm, RELU &ConvTranspose2d(256,1,3,2,padding = 1),Sigmoid \\
        Linear(d), Linear(1024,d)& \\

        \midrule
        \headcol Trajectory Encoder & Trajectory Decoder \\
        \midrule
        Linear(512), Batchnorm, LeakyRELU&  Linear(512), Batchnorm, LeakyRELU\\
        Linear(512), Batchnorm, LeakyRELU& Linear(512), Batchnorm, LeakyRELU\\
        Linear(512), Batchnorm, LeakyRELU& Linear(512), Batchnorm, LeakyRELU\\
        Linear(d), Linear(d)& Linear(128), Sigmoid\\
        
        \midrule
    \headcol Sound Encoder & Sound Decoder \\
    \midrule
    Conv2d(1,128, kernel = (1,128), stride = (1,1)), Batchnorm, RELU & Linear(2048), Batchnorm, RELU \\
    Conv2d(128,128, kernel=(4,1), stride = (2,1)), Batchnorm, RELU & ConvTranspose2d(256, 128, kernel=(4,1), stride=(2,1), padding=(1,0))\\
    Conv2d(128,256, kernel=(4,1),stride = (2,1)), Batchnorm, RELU &ConvTranspose2d(128, 128, kernel=(4,1), stride=(2,1), padding=(1,0))\\
    Linear(d), Linear(d)&ConvTranspose2d(128, 1, kernel=(1,128), stride=(1,1), padding=0), Sigmoid\\
    \midrule
    \headcol Training parameters &  Joint Encoder \\
    \midrule
    Batchsize = 64 & Separates head for each modality with same architectures as encoders\\
    Learning rate = 1e-3 & Linear(512), RELU\\
    Optimizer = Adam & Linear(512), RELU \\
     & Linear(d), Linear(d) \\
    $\lambda_{image} = \frac{32\times128}{3\times28\times28}\approx 1.7 $ &\\
    $\lambda_{audio} = 1.0$& \\
    $\lambda_{trajectory} = \frac{32\times128}{200} = 20.48$&\\
    \midrule
    \headcol Normalizing Flows & JNF-Shared encoders for $\qgj$\\
    \midrule
    Masked Autoregressive with two MADE blocks  & Linear(10,512) RELU\\
                                                & Linear(512,512) RELU \\
                                                & Linear(512,20), Linear(512,20)\\
    \bottomrule
\end{tabular}
\caption{Architectures and training parameters used for the MHD Experiments.}\label{tab:mhd parameters}

\end{sidewaystable}
\newpage
\section{Hamiltonian Monte Carlo Sampling}
\label{app:hmc}
In this appendix, we recall the principles of Hamiltonian Monte Carlo Sampling and detail how we apply it in our model. 
The Hamiltonian Monte Carlo (HMC) sampling belongs to the larger class of Markov Chain Monte Carlo methods (MCMC) that allow to sample from any distribution $f(z)$ known up to a constant \cite{noauthor_mcmc_2011}. 
The general principle is to build a Markov Chain that will have our target $f(z)$ as stationary distribution. More specifically, the HMC is an instance of the Metropolis-Hasting Algorithm (see \ref{alg:metropolis}) that uses a physics-oriented proposal distribution. 
\begin{algorithm}[htb]
   \caption{Metropolis-Hasting Algorithm}
   \label{alg:metropolis}
\begin{algorithmic}[1]
   \State {\bfseries Initialization : $z \gets z_0$}
   \For {$i \coloneqq 0$ $\rightarrow$ N} 
    \State {Sample $z'$ from the proposal $g(z'|z)$}
   \State {With probability $\alpha(z',z)$ accept the proposal $z\gets z'$}
   \EndFor
\end{algorithmic}
\end{algorithm}

Sampling from the proposal distribution $g(z'|z_0)$ is done by integrating the Hamiltonian equations : 
\begin{equation}
    \label{HMC-evol}
    \left \{ \begin{aligned}
        & \frac{\partial z}{\partial t} = \frac{\partial H}{\partial v}\,,\\
        & \frac{\partial v}{\partial t} = -\frac{\partial H}{\partial z}\,,\\
        & z(0) = z_0\,\\
        & v(0) = v_0 \sim \mathcal{N}(0,I)\,,\\
    \end{aligned} \right .
\end{equation}
where the Hamiltonian is defined by $H(z,v) = - \log f(z) + \frac{1}{2}v^tv $. 
In physics, Eq.~\eqref{HMC-evol} describes the evolution in time of a physical particle with initial position $z$ and a random initial momentum $v$. The leap-frog numerical scheme is used to integrate Eq.~\eqref{HMC-evol} and is repeated $l$ times with a small integrator step size $\epsilon$ :
\begin{equation}
    \label{leapfrog}
    \begin{split}
        &v(t + \frac{\epsilon}{2}) = v(t) + \frac{\epsilon}{2} \cdot \nabla_z(\log f(z)(t))\,,\\
        &z(t+\epsilon) = z(t) + \epsilon \cdot v(t +\frac{\epsilon}{2})\,,\\
        &v(t+ \epsilon) = v(t+\frac{\epsilon}{2}) + \frac{\epsilon}{2} \nabla_z \log f(z(t+\epsilon))\,.\\
    \end{split}
\end{equation}
After $l$ integration steps, we obtain the proposal position $z'= z(t + l\cdot\epsilon)$ that corresponds to step $3$ in Algorithm \ref{alg:metropolis}. The acceptance ratio is then defined as $\alpha(z',z_0) = \min\left(1, \frac{\exp(-H(z_0,v_0))}{\exp(-H(z',v(t + l\cdot\epsilon)))}\right)$. 
This procedure is repeated to produce an ergodic Markov chain $(z^n)$ converging to the target distribution $f$ \cite{duane_hybrid_1987, liu_monte_2009, neal_mcmc_2011, girolami_riemann_2011}. In this work, we use HMC sampling to sample from the PoE of unimodal posteriors in Eq.~\eqref{eq:poe}. To do so we need to compute and derivate the (log) of the target distribution given by the PoE of the unimodal distributions:
\begin{equation}
\log q(z|(x_i)_{i\in S}) = -\log p(z) + \sum_{i \in S} \log q_{\phi_i}(z|x_i)\,.
\end{equation}
We can use autograd to automatically compute the gradient $\nabla_z \log q(z|(x_i)_{i\in S})$ that is needed in the leapfrog steps. 

In our experiments, we use 100 steps per sampling. 

\section{Information on the classifiers used for evaluation}

\subsection{MNIST-SVHN}

In Table~\ref{classifiers ms} we provide the architectures and the accuracies for the classifiers that we use to evaluate coherence on the MNIST-SVHN dataset. 

\begin{table}[H]
    \footnotesize
    \centering
    \begin{tabular}{@{}ll@{}}
        \toprule
        \bf{SVHN} & \bf{MNIST}\\
        \midrule
        Conv2d(3,10,5) & Conv2d(1,10,5) \\
         MaxPool2d,RELU & MaxPool2d,RELU \\

        Conv2d(10,20,5), Dropout(0.5) & Conv2d(10,20,5), Dropout(0.5)  \\
         MaxPool2d,RELU &  MaxPool2d,RELU  \\
        Linear(500,50), RELU, Dropout & Linear(350,50), RELU, Dropout \\
        Linear(50,10), Softmax & Linear(50,10), Softmax \\
        \midrule
        \multicolumn{2}{c}{\bf{Accuracies on test}}\\
        \midrule
        0.87 & 0.99\\
    \end{tabular}
\caption{Classifiers used for the MNIST-SVHN experiments.}\label{classifiers ms}
\end{table}

\subsection{Classifiers on PolyMNIST}

We use the architectures and the pretrained models available at \url{https://github.com/thomassutter/MoPoE}  \cite{sutter_generalized_2021}.

The accuracies of the classifiers for the five modalities of the test set are respectively: 0.95, 0.99, 0.99, 0.97, 0.95. 

\subsection{Classifiers on TranslatedPolyMNIST}

We pretrain classifiers on this dataset having similar architectures as in Figure~\ref{fig:resnets} with a output size of 10. 

The accuracies of the trained classifiers for the five modalities of the test set are respectively: 0.98, 0.97, 0.98, 0.97, 0.98. 

\subsection{Classifiers on MHD}

We use the pretrained classifiers available at \url{https://github.com/miguelsvasco/nexus_pytorch/}. 

The accuracies of the trained classifiers on the test set are: 0.95 for the audio modality, 0.99 for the image modality and 0.99 for the trajectory modality.

\bibliographystyle{elsarticle-num} 
\bibliography{biblio}{}

\end{document}